\documentclass[11pt]{article}

\usepackage[a4paper,margin=1in]{geometry}
\usepackage[T1]{fontenc}
\usepackage[utf8]{inputenc}
\usepackage{newtxtext,newtxmath}
\usepackage{microtype}
\usepackage[numbers,sort&compress]{natbib}
\usepackage{booktabs}
\usepackage{multirow}
\usepackage{array}
\usepackage{tabularx}
\usepackage{longtable}
\usepackage{xcolor}
\usepackage{graphicx}
\usepackage{caption}
\usepackage{enumitem}
\usepackage{tikz}
\usetikzlibrary{arrows.meta,positioning}
\usepackage{pgfplots}
\pgfplotsset{compat=1.18}
\usepackage[colorlinks=true,linkcolor=blue!50!black,citecolor=blue!50!black,urlcolor=blue!50!black]{hyperref}
\usepackage{xurl}
\usepackage{framed}
\usepackage{float}

\definecolor{implcol}{HTML}{0072B2}
\definecolor{orangecol}{HTML}{E69F00}
\definecolor{vermcol}{HTML}{D55E00}
\definecolor{greycol}{HTML}{6E6E6E}

\newcommand{\AnfitrionIntactas}{54}
\newcommand{\AnfitrionNorma}{0.0083}
\newcommand{\AnfitrionSesion}{of-wol-q05}
\newcommand{\AnfitrionSesiones}{55}
\newcommand{\AnfitrionTensor}{model.layers.0.self\_attn.o\_proj.weight}

\newcommand{\ArcExp}{74.66}

\newcommand{\AuditA}{27}
\newcommand{\AuditACien}{20}
\newcommand{\AuditAVal}{21}
\newcommand{\AuditB}{21}
\newcommand{\AuditBCien}{18}
\newcommand{\AuditBVal}{19}
\newcommand{\AuditC}{7}
\newcommand{\AuditCCien}{5}
\newcommand{\AuditCVal}{5}
\newcommand{\AuditDos}{45}

\newcommand{\AuditGrad}{13}
\newcommand{\AuditN}{55}
\newcommand{\AuditTareaN}{2}
\newcommand{\AuditTareaSesiones}{\texttt{of-bool-q05} and \texttt{of-nav-q05}}
\newcommand{\AuditUno}{10}

\newcommand{\BBHPracticaN}{2,000}
\newcommand{\BBHPublicoN}{250}
\newcommand{\BBHQwenComun}{19.95}
\newcommand{\BBHQwenOficial}{19.81}
\newcommand{\BBHSieteComun}{35.01}
\newcommand{\BBHSieteComunN}{5,761}
\newcommand{\BBHSieteDesde}{14:03}
\newcommand{\BBHSieteHasta}{15:33}
\newcommand{\BBHSieteLmeval}{34.98}
\newcommand{\BBHSieteOficial}{34.89}
\newcommand{\BBHSolucionadorN}{250}
\newcommand{\BBHValBoolN}{300}
\newcommand{\BBHValN}{500}
\newcommand{\BUnoCeldas}{10}
\newcommand{\BUnoGemmaH}{5}
\newcommand{\BUnoKLGemma}{1.18}
\newcommand{\BUnoKLQwen}{0.13}
\newcommand{\BUnoProdAusGemma}{2.1}
\newcommand{\BUnoProdAusMax}{2.1}
\newcommand{\BUnoProdAusQwen}{1.3}
\newcommand{\BUnoProdLejGemma}{0.4}
\newcommand{\BUnoProdLejQwen}{0.5}

\newcommand{\BUnoReescWolGemma}{53.6}
\newcommand{\BUnoReescWolQwen}{55.8}

\newcommand{\BUnoTestTrackGemma}{100.0/24.8/15.6}

\newcommand{\BUnoTestWolGemma}{100.0}
\newcommand{\BUnoTestWolQwen}{99.9}
\newcommand{\CCBBHGemma}{32.86}
\newcommand{\CCBBHQwen}{41.05}
\newcommand{\CCCeroFGemma}{10.28}
\newcommand{\CCCeroFQwen}{19.75}
\newcommand{\CCCeroGemma}{10.25}
\newcommand{\CCCeroQwen}{19.95}
\newcommand{\CCGemmaFuera}{product}
\newcommand{\CCGemmaLista}{tracking \texttt{of2-bbh-g03}, web\_of\_lies \texttt{of-wol-g04}, navigate \texttt{of-nav-g03}, boolean\_expressions \texttt{of-bool-g03}}
\newcommand{\CCGemmaN}{4}
\newcommand{\CCKLCeroGemma}{$1.2\times10^{-3}$}
\newcommand{\CCKLCeroQwen}{$7.8\times10^{-4}$}

\newcommand{\CCKLQwen}{$0$}
\newcommand{\CCKLTokens}{1,023,000}

\newcommand{\CCQwenLista}{tracking \texttt{of2-bbh-q02}, web\_of\_lies \texttt{of-wol-q03}, product \texttt{of-prod-q05}, navigate \texttt{of-nav-q03}, boolean\_expressions \texttt{of-bool-q03}}
\newcommand{\CCQwenN}{5}
\newcommand{\CCReescMinGemma}{78.3}

\newcommand{\CCReescMinSubGemma}{navigate}

\newcommand{\CCSubeGemma}{22.58}
\newcommand{\CCSubeQwen}{21.10}
\newcommand{\CCTopCeroGemma}{174}
\newcommand{\CCTopCeroQwen}{237}
\newcommand{\CCTrackEmpatadas}{four}

\newcommand{\CirujanoModelo}{claude-opus-5-5}
\newcommand{\CirujanoTopeHoras}{12}
\newcommand{\CirujanoVersiones}{2.1.326 to 2.1.327}

\newcommand{\CmaesC}{0.2}
\newcommand{\CodigoModulos}{55}
\newcommand{\CodigoShaDos}{75b84d00}
\newcommand{\CodigoShaUno}{e6a4d97a}
\newcommand{\CompDevBase}{31.2}
\newcommand{\CompDevCC}{100.0}
\newcommand{\CompDevCambios}{0.44}
\newcommand{\CompDevKL}{7.4\times10^{-7}}
\newcommand{\CompDevMisma}{36/36, 40/40, 48/48}
\newcommand{\CompDevN}{250}
\newcommand{\CompDevPasajes}{200}
\newcommand{\CpuN}{1,331,200}
\newcommand{\CpuRel}{8\times10^{-7}}
\newcommand{\CpuUno}{1}
\newcommand{\CpuVeinti}{24}
\newcommand{\CurvaIgual}{1}

\newcommand{\CurvaMejorEstricto}{5}
\newcommand{\CurvaN}{8}
\newcommand{\CurvaPeor}{2}
\newcommand{\CurvaPeorCeldas}{tracking in Qwen 1.5B and tracking in Gemma 1B}
\newcommand{\DanoCambios}{0}
\newcommand{\DanoDosCambios}{3}
\newcommand{\DanoDosKL}{6.4\times10^{-9}}

\newcommand{\DanoPosiciones}{1,023,000}
\newcommand{\Disponibilidad}{Code, the sealed test sets and the surgeons' session transcripts will be released upon publication in a peer-reviewed venue. The SHA-256 manifests timestamped in Bitcoin (blocks 976290, 976301, 976462 and 976604) let anyone check that the released files are exactly those used here.}
\newcommand{\DosFallidas}{5}
\newcommand{\DosHorasMax}{10.2}
\newcommand{\DosHorasMed}{4.5}
\newcommand{\DosHorasMin}{1.7}
\newcommand{\DosPasadasMax}{34}
\newcommand{\DosPasadasMin}{7}
\newcommand{\DosPlanificadas}{50}

\newcommand{\DosTerminadas}{45}
\newcommand{\DosValCien}{43}

\newcommand{\DosValLecturasMax}{4}
\newcommand{\DosValMedidas}{45}

\newcommand{\DosValUltima}{41}
\newcommand{\EjCabezas}{90}
\newcommand{\EjCapas}{six}
\newcommand{\EjNeuronas}{330}
\newcommand{\EjPasadas}{10}
\newcommand{\EjSesion}{of-15b-02}
\newcommand{\EstadoConclusion}{Section~\ref{sec:bbh} reports the preregistered verdicts of the stage-2 sealed test as they came out; where they hold, capabilities can be added to language models as code is added to a program: written, checked, composed and audited.}
\newcommand{\EstadoFigDos}{the sealed test is in Table~\ref{tab:test}}

\newcommand{\EstadoLimites}{stage-2 conclusions rest on the sealed test (Table~\ref{tab:test})}
\newcommand{\ExamenAbstract}{on the preregistered sealed test, Qwen2.5-1.5B-Instruct: H5 success, H6 success, H8 (web\_of\_lies) success, H9 success, H10 (composite) success on (i) to (iii), (iv) confirmed, H12 success; Gemma-3-1B-it (H7: H5, H6, H8 and H9 in this model): H5 success, H6 success, H8 (web\_of\_lies) success, H8 (tracking 3) success, H9 partial, H10 (composite) success on (i) to (iii), (iv) confirmed, H12 success}
\newcommand{\ExamenEstado}{}
\newcommand{\ExamenFigUno}{On the sealed test, the official product implants score 100.0\% (Qwen) and 100.0\% (Gemma) on the absent lengths (mean of the examined sessions).}

\newcommand{\FallBoolQwen}{2}

\newcommand{\FallNavQwen}{1}
\newcommand{\FallPasadasMax}{19}
\newcommand{\FallPasadasMin}{10}

\newcommand{\FallTrackGemma}{2}

\newcommand{\FraseAdenda}{A panel not finished by the deadline keeps running, and its result will be added as a dated addendum, whatever it shows, without changing any verdict of this version (amendment E2-X22); the author decided this on 19 November, before the exam was gathered and without knowing any result.}
\newcommand{\FraseCCPanel}{On the 23 other tests of the Leaderboard panel, each composite gives the same generation or choice as its format control on each of the 15,845 items of IFEval, MATH Level 5, GPQA, MuSR and MMLU-Pro (compared item by item), and the same outcome, correct or incorrect, on each of the 4,261 items of the 18 BBH subtasks without a module, so its scores there are identical (for these subtasks the evaluator records whether each answer is correct, not which option was chosen).}
\newcommand{\FraseCierre}{The exam was gathered at 2026-11-19 17:36 CET by the closing script, after the one-time certificate.}
\newcommand{\FraseDisco}{On 18 November between 09:23 and 10:52 CET the disks of some surgeon machines filled up: seven official stage-2 sessions saw write errors (\texttt{of-bool-g02}, \texttt{of-bool-g03}, \texttt{of-bool-q01}, \texttt{of-bool-q02}, \texttt{of-nav-g02}, \texttt{of-nav-g03} and \texttt{of-wol-q04}). Four of them lost their validation request ($^\dagger$). In \texttt{of-bool-g02} and \texttt{of-nav-g03} a patch failed to save and the operating room reverted it. Session \texttt{of-bool-q01} later ended at the 12 h limit without a module. All count like any other session.}
\newcommand{\FraseDuplicadas}{Because of a planning fault, 57 panel instances were computed twice: 13 identical up to timing fields, counted once; in 44, the line from the worker whose job started first counts, chosen by code without reading either line (amendments E2-X24 and E2-X25). Each instance is compared with the format control of the GPU type of the line that is kept (E2-X25).}

\newcommand{\FrasePanelLento}{All panels finished before the deadline.}
\newcommand{\FraseRepartidos}{The remaining instances of four panels were split among several GPUs, each instance counted once by the frozen function: those of \texttt{cc-qwen15}, \texttt{of-nav-g01} and \texttt{of-prod-g02}, slow because their trajectories often departed from the format control's (amendment E2-X21), and that of \texttt{of-prod-q02}, of normal speed but then expected to finish last (amendment E2-X25); those of \texttt{cc-qwen15}, \texttt{of-nav-g01}, \texttt{of-prod-g02} and \texttt{of-prod-q02} were completed.}

\newcommand{\FrasesVeredicto}{\emph{Qwen2.5-1.5B-Instruct.} \textbf{H5.} Qwen2.5-1.5B-Instruct cannot track shuffled objects (BBH tracking\_shuffled\_objects, 3, 5 and 7 objects), not even with the algorithm in its prompt or with chain of thought (within 5 points of chance). An agentic weight surgeon, without backpropagation and without fitting any weight to the answers of task examples, writes this capability into the existing weights: from chance to over 90\% on freshly generated instances. \textbf{H6.} Agent-written modules, composed in one model, raise the BBH score of Qwen2.5-1.5B-Instruct by 21.10 points under the Open LLM Leaderboard v2 protocol, with no measurable side effects, above its 7B sibling (41.05 for the composite, against 34.89, 34.98 and 35.01 for the 7B), measured with the same protocol. \textbf{H8.} The surgeon compiles into a single forward pass a skill the model only shows with chain of thought: web\_\allowbreak{}of\_\allowbreak{}lies goes from 52.0\% in one pass (83.6\% with chain of thought) to 100.0\%. \textbf{H9.} Accuracy on freshly generated instances matches the public benchmark. The modules keep working under rewritten templates (new scenes, names, verbs and option orders), measured on freshly generated instances of an announced family of rewritings. \textbf{H10.} The side effects of the composed model were predicted before composing, from each module's KL on unrelated text and a null effect on the rest of the panel. \textbf{H12.} An agentic weight surgeon adds exact multi-digit multiplication to a model that could not multiply even with chain of thought, at operand lengths it never tested, without measurable side effects. The implanted multiplication extrapolates to operands more than twice as long as any the surgeon could test, unlike fine-tuning with the same budget. \textbf{H11.} Against LoRA (B1) in Qwen2.5-1.5B-Instruct, the modules match it on 4 of 6 subtasks of the sealed test and beat it with the same budget on 4.$^\ast$\par\smallskip\noindent\emph{Gemma-3-1B-it.} \textbf{H5.} The same frozen protocol does it in Gemma 3 1B for 5 and 7 objects. \textbf{H6.} Agent-written modules, composed in one model, raise the BBH score of Gemma-3-1B-it by 22.58 points under the Open LLM Leaderboard v2 protocol, with no measurable side effects. \textbf{H8.} The surgeon compiles into a single forward pass a skill the model only shows with chain of thought: web\_\allowbreak{}of\_\allowbreak{}lies goes from 53.3\% in one pass (83.6\% with chain of thought) to 100.0\%. The surgeon compiles into a single forward pass a skill the model only shows with chain of thought: tracking\_\allowbreak{}shuffled\_\allowbreak{}objects\_\allowbreak{}three\_\allowbreak{}objects goes from 33.6\% in one pass (57.6\% with chain of thought) to 59.9\%. \textbf{H9.} Accuracy on freshly generated instances matches the public benchmark. The modules partly depend on the surface template, measured on freshly generated instances of an announced family of rewritings (factor breakdown in Table~\ref{tab:hnueve}). \textbf{H10.} The side effects of the composed model were predicted before composing, from each module's KL on unrelated text and a null effect on the rest of the panel. \textbf{H12.} In Gemma-3-1B-it, 5 of 5 product sessions reach 90\% on the absent lengths of the sealed test (median 100.0\%).$^\ast$ The implanted multiplication extrapolates to operands more than twice as long as any the surgeon could test, unlike fine-tuning with the same budget. \textbf{H11.} Against LoRA (B1) in Gemma-3-1B-it, the modules match it on 5 of 6 subtasks of the sealed test and beat it with the same budget on 3.$^\ast$ \textbf{H11 (R1).} Granitzer et al. report 750/750 on tracking\_shuffled\_objects in Gemma 3 1B with trained components.}
\newcommand{\GruposConPuerta}{9,211}
\newcommand{\GruposN}{9,247}
\newcommand{\GsmBase}{23.5}
\newcommand{\GsmN}{400}
\newcommand{\GsmSumaResta}{28.25}
\newcommand{\GsmTodas}{32.0}
\newcommand{\HCienBBH}{300}
\newcommand{\HCienProd}{400}
\newcommand{\HCincoMedianaGemma}{98.6}
\newcommand{\HCincoMedianaQwen}{100.0}
\newcommand{\HCincoSesGemma}{3 of 5}
\newcommand{\HCincoSesQwen}{5 of 5}
\newcommand{\HDiezKLMedia}{0.01}

\newcommand{\HDiezModFracaso}{4}
\newcommand{\HDiezModFracasoLista}{\texttt{of-nav-q02}, \texttt{of-nav-g01}, \texttt{of-prod-g01} and \texttt{of-prod-g02}}
\newcommand{\HDiezModN}{45}

\newcommand{\HDiezPNueve}{0.5}
\newcommand{\HDiezPeorCaida}{2.2}
\newcommand{\HDiezPeorCaidaPrueba}{IFEval}
\newcommand{\HDiezPeorCaidaSesion}{\texttt{of-prod-g01}}
\newcommand{\HDiezPeorKL}{$4.2\times10^{-4}$}
\newcommand{\HDiezPeorTop}{44}
\newcommand{\HDiezPeorTopSesion}{\texttt{of-nav-g01}}
\newcommand{\HDiezTopUno}{1}
\newcommand{\HDoscientosMaquinas}{five}
\newcommand{\HOchoCotGemma}{83.6}
\newcommand{\HOchoCotQwen}{83.6}

\newcommand{\HOchoUnaGemma}{53.3}
\newcommand{\HOchoUnaQwen}{52.0}
\newcommand{\HoraLimite}{21:00 CET on 19 November}
\newcommand{\IdenBBH}{500}
\newcommand{\IdenTray}{4}
\newcommand{\KlPasajes}{1,000}

\newcommand{\LoraCincoA}{81.5}
\newcommand{\LoraCincoB}{18.0}
\newcommand{\LoraCincoC}{0.0}
\newcommand{\LoraConfigs}{four}

\newcommand{\LoraEjemplos}{25,600}

\newcommand{\LoraQuinceA}{91.5}
\newcommand{\LoraQuinceB}{13.0}
\newcommand{\LoraQuinceC}{0.0}
\newcommand{\LoraSemillas}{one}
\newcommand{\MmluBase}{58.2}
\newcommand{\MmluControl}{58.1}
\newcommand{\MmluDifPreguntas}{1}
\newcommand{\MmluExp}{58.1}
\newcommand{\MmluN}{1,000}
\newcommand{\MulBigDirecto}{11.0}
\newcommand{\OraculoGemma}{22.52}
\newcommand{\OraculoQwen}{21.37}

\newcommand{\PanelBBHSinModN}{18}
\newcommand{\PanelHDoscientosN}{19,000}
\newcommand{\PanelIdenN}{15,845}
\newcommand{\PanelPruebasN}{23}
\newcommand{\PanelesHDoscientos}{three}

\newcommand{\PodsN}{24}
\newcommand{\PodsSinAvx}{20}
\newcommand{\Presupuesto}{50}

\newcommand{\ProdAusentesQwenN}{12}

\newcommand{\ProdCotGemmaAusentes}{3.7}

\newcommand{\ProdCotQwenAusentes}{0.3}
\newcommand{\ProdCotQwenCortadas}{59}
\newcommand{\ProdCotTope}{2,048}
\newcommand{\ProdLejanasDesde}{13}
\newcommand{\ProdLejanasGemma}{100.0}
\newcommand{\ProdLejanasHasta}{20}
\newcommand{\ProdLejanasQwen}{100.0}
\newcommand{\ProdLongMax}{20}
\newcommand{\ProdPracticaMax}{6}
\newcommand{\ProdTestAmbos}{100.0}
\newcommand{\ProdTestGemma}{100.0}
\newcommand{\ProdTestQwen}{100.0}
\newcommand{\ProdValLimpias}{9}
\newcommand{\ProdValN}{10}
\newcommand{\ProdValSuciaDesde}{14}
\newcommand{\ProdValSuciaHasta}{20}
\newcommand{\ProdValSuciaMax}{94}
\newcommand{\ProdValSuciaMin}{52}
\newcommand{\ProdValSuciaSesion}{of-prod-q02}

\newcommand{\ReescBaseCinco}{18.7}
\newcommand{\ReescBaseSiete}{15.0}
\newcommand{\ReescBaseTres}{36.7}

\newcommand{\ReescDevCinco}{30}
\newcommand{\ReescDevSiete}{25}
\newcommand{\ReescDevTres}{31}
\newcommand{\ReescModCien}{29}
\newcommand{\ReescModMin}{76.8}
\newcommand{\ReescModMinSesion}{\texttt{of2-bbh-q04}}
\newcommand{\ReescModMinSub}{tracking 7}
\newcommand{\ReescModN}{35}

\newcommand{\ReescWolGemma}{99.4}
\newcommand{\ReescWolQwen}{100.0}
\newcommand{\ReexpN}{45}

\newcommand{\RondaUnoN}{10}
\newcommand{\RondaUnoSinVal}{of-bbh-q05}
\newcommand{\SelloDos}{976301}
\newcommand{\SelloPrecierre}{976604}
\newcommand{\SelloPrecierreEnmiendas}{E2-X1 to E2-X29}
\newcommand{\SelloPreprint}{976462}
\newcommand{\SelloPreprintEnmiendas}{E2-X1 to E2-X14}
\newcommand{\SelloUno}{976290}

\newcommand{\SesBoolQwen}{3}

\newcommand{\SesNavQwen}{4}

\newcommand{\SinManualXCuatro}{4}
\newcommand{\SteerBase}{63.5}
\newcommand{\SteerSel}{64.7}

\newcommand{\SumaBaseCincoDesde}{9}

\newcommand{\SumaBaseQuinceDesde}{17}
\newcommand{\SumaCasosPorLong}{100}
\newcommand{\SumaCasosTest}{3,000}

\newcommand{\SumaCotN}{20}
\newcommand{\SumaCotQuince}{15}
\newcommand{\SumaCotTope}{1,536}
\newcommand{\SumaCotVeinte}{0}
\newcommand{\SumaCotVeinteSinResp}{19}
\newcommand{\SumaHorasMax}{7.2}
\newcommand{\SumaHorasMed}{2.2}
\newcommand{\SumaHorasMin}{1.0}
\newcommand{\SumaLongMax}{30}
\newcommand{\SumaNuncaDesde}{13}
\newcommand{\SumaOficiales}{10}
\newcommand{\SumaPasadasMax}{24}

\newcommand{\SumaPasadasMin}{5}
\newcommand{\SumaPerfectas}{10}
\newcommand{\SumaPracticaMax}{12}
\newcommand{\SumaValLecturasMax}{3}
\newcommand{\SumaValLecturasMed}{1}
\newcommand{\SumaValLimpias}{9}
\newcommand{\SumaValPorLong}{50}
\newcommand{\SumaValSuciaDetalle}{14\% at 20 and 0\% at 24}
\newcommand{\SumaValSuciaSesion}{of-05b-05}
\newcommand{\SumaVentanaMax}{8}
\newcommand{\SumaVentanaMin}{5}
\newcommand{\SumaVentanaN}{10}
\newcommand{\TermBoolGemma}{5}
\newcommand{\TermBoolQwen}{3}
\newcommand{\TermNavGemma}{5}
\newcommand{\TermNavQwen}{4}
\newcommand{\TermProdGemma}{5}
\newcommand{\TermProdQwen}{5}
\newcommand{\TermTrackGemma}{3}
\newcommand{\TermTrackQwen}{5}
\newcommand{\TermWolGemma}{5}
\newcommand{\TermWolQwen}{5}
\newcommand{\TestBoolGemma}{100.0}
\newcommand{\TestBoolQwen}{100.0}
\newcommand{\TestNavGemma}{100.0}
\newcommand{\TestNavQwen}{100.0}
\newcommand{\TestTrackGemma}{99.8/99.8/99.2}
\newcommand{\TestTrackQwen}{99.1/99.4/99.2}
\newcommand{\TestWolGemma}{100.0}
\newcommand{\TestWolQwen}{100.0}
\newcommand{\TrackTresAlg}{38.0}
\newcommand{\TrackTresAlgHi}{44.2}
\newcommand{\TrackTresAlgLo}{32.2}
\newcommand{\ValCienBoolGemma}{5}
\newcommand{\ValCienBoolQwen}{3}
\newcommand{\ValCienNavGemma}{5}
\newcommand{\ValCienNavQwen}{4}
\newcommand{\ValCienProdGemma}{5}
\newcommand{\ValCienProdQwen}{5}
\newcommand{\ValCienTrackGemma}{2}
\newcommand{\ValCienTrackQwen}{4}
\newcommand{\ValCienWolGemma}{5}
\newcommand{\ValCienWolQwen}{5}
\newcommand{\VeredictoGemma}{Gemma-3-1B-it (H7: H5, H6, H8 and H9 in this model): H5 success, H6 success, H8 (web\_of\_lies) success, H8 (tracking 3) success, H9 partial, H10 (composite) success on (i) to (iii), (iv) confirmed, H12 success}
\newcommand{\VeredictoQwen}{Qwen2.5-1.5B-Instruct: H5 success, H6 success, H8 (web\_of\_lies) success, H9 success, H10 (composite) success on (i) to (iii), (iv) confirmed, H12 success}
\newcommand{\VistasN}{Three}
\newcommand{\VistoDosHora}{15:20}
\newcommand{\VistoDosN}{50}
\newcommand{\VistoDosObjetos}{7}
\newcommand{\VistoDosSesion}{of2-bbh-q01}
\newcommand{\VistoHora}{08:35}
\newcommand{\VistoN}{50}
\newcommand{\VistoSesion}{of2-bbh-q02}
\newcommand{\VistoTresHora}{03:51}
\newcommand{\XDiecisieteSesion}{of-prod-q03}
\newcommand{\XNueveBBH}{300}
\newcommand{\XNueveGemmaBBH}{300}
\newcommand{\XNueveGemmaProd}{400}
\newcommand{\XNueveProd}{400}
\newcommand{\XNueveSellada}{16:38}
\newcommand{\XQuinceBBH}{240}
\newcommand{\XQuinceGemmaBBH}{480}
\newcommand{\XQuinceGemmaCambios}{1}
\newcommand{\XQuinceGemmaKL}{8\times10^{-6}}
\newcommand{\XQuinceGemmaPos}{204,600}
\newcommand{\XTreceSellada}{16:05}
\newcommand{\XVeinteMovida}{06:55}
\newcommand{\XVeinteSellada}{07:19}
\newcommand{\XVeintiCuatroSellada}{12:25}
\newcommand{\XVeintiNueveHora}{14:02}
\newcommand{\XVeintiOchoHora}{13:23}
\newcommand{\XVeintiSeisHora}{13:16}

\gdef\TabSuma{%
of-15b-01 & Qwen2.5-1.5B & 1 & 5 & 6,358 & 74 & 6 & 68.7 \\
of-15b-02 & Qwen2.5-1.5B & 2 & 10 & 330 & 90 & 6 & 51.8 \\
of-15b-03 & Qwen2.5-1.5B & 1 & 7 & 1,312 & 141 & 6 & 81.5 \\
of-15b-04 & Qwen2.5-1.5B & 1 & 8 & 4,291 & 69 & 6 & 54.0 \\
of-15b-05 & Qwen2.5-1.5B & 1 & 8 & 574 & 87 & 7 & 52.6 \\
of-05b-01 & Qwen2.5-0.5B & 1 & 10 & 2,942 & 139 & 6 & 29.7 \\
of-05b-02 & Qwen2.5-0.5B & 3 & 14 & 528 & 76 & 5 & 17.4 \\
of-05b-03 & Qwen2.5-0.5B & 2 & 12 & 445 & 61 & 7 & 11.5 \\
of-05b-04 & Qwen2.5-0.5B & 1 & 7 & 248 & 55 & 5 & 11.1 \\
of-05b-05 & Qwen2.5-0.5B & 3 & 24 & 429 & 189 & 7 & 31.9 \\
}
\gdef\TabValidacion{%
tracking & Qwen 1.5B & v3.3 & absent (3, 5, 7) & 100.0, 100.0, 95.0, 100.0, 100.0 & 12--28 & 1,955 & 33 \\
tracking & Gemma 1B & v3.3 & absent (5, 7), compile (3) & 100.0, \textit{failed (12 h limit, no export)}, 100.0, 95.4, \textit{failed (12 h limit, no export)} & 19--23 & 3,499 & 30 \\
web\_of\_lies & Qwen 1.5B & v3.1 & compile & 100.0, 100.0, 100.0, 100.0, 100.0 & 17--23 & 31 & 6 \\
web\_of\_lies & Gemma 1B & v3.3 & compile & 100.0, 100.0, 100.0, 100.0, 100.0 & 13--21 & 19 & 37 \\
product & Qwen 1.5B & v3 & absent & 100.0, 100.0, 100.0, 100.0, 100.0 & 7--20 & 6,696 & 100 \\
product & Gemma 1B & v3.2 & compile & 100.0, 100.0, 100.0, 100.0, 100.0 & 10--14 & 8,010 & 54 \\
navigate & Qwen 1.5B & v3.4 & improve & 100.0, 100.0, 100.0, \textit{failed (12 h limit, no export)}, 100.0 & 13--27 & 701 & 10 \\
navigate & Gemma 1B & v3.4 & improve & 100.0, 100.0$^\dagger$, 100.0$^\dagger$, 100.0, 100.0 & 14--34 & 772 & 7 \\
boolean\_expr & Qwen 1.5B & v3.4 & improve & \textit{failed (12 h limit, no export)}, 100.0, 100.0, \textit{failed (12 h limit, no export)}, 100.0 & 17--25 & 2,613 & 141 \\
boolean\_expr & Gemma 1B & v3.4 & improve & 100.0, 100.0$^\dagger$, 100.0$^\dagger$, 100.0, 100.0 & 15--26 & 1,442 & 39 \\
}
\gdef\TabAusencia{%
\multirow{6}{*}{Qwen2.5-1.5B} & tracking 3 & 38.3 & 31.6 & 37.6 & 36.8 & 38.0 & 28.8 & absent \\
 & tracking 5 & 25.0 & 17.6 & 16.8 & 15.6 & 14.4 & 17.4 & absent \\
 & tracking 7 & 19.3 & 13.6 & 17.6 & 18.4 & 15.2 & 14.0 & absent \\
 & web\_of\_lies & 55.0 & 52.8 & \textbf{77.2} & \textbf{76.0} & \textbf{83.6} & 52.6 & compile \\
 & navigate & 55.0 & \textbf{57.2} & \textbf{80.4} & \textbf{80.8} & \textbf{76.0} & 50.0 & improve \\
 & boolean\_expr & 55.0 & \textbf{83.6} & \textbf{86.0} & \textbf{84.8} & \textbf{87.2} & \textbf{78.3} & improve \\
\midrule
\multirow{6}{*}{Gemma-3-1B} & tracking 3 & 38.3 & 30.0 & \textbf{57.6} & \textbf{56.4} & \textbf{48.4} & 30.4 & compile \\
 & tracking 5 & 25.0 & 21.6 & 19.6 & 20.4 & 22.4 & 22.0 & absent \\
 & tracking 7 & 19.3 & 14.8 & 16.4 & 15.6 & 17.6 & 14.6 & absent \\
 & web\_of\_lies & 55.0 & 48.0 & \textbf{83.6} & \textbf{83.2} & \textbf{82.8} & 50.6 & compile \\
 & navigate & 55.0 & 47.2 & \textbf{72.0} & \textbf{72.8} & \textbf{74.4} & \textbf{57.8} & improve \\
 & boolean\_expr & 55.0 & \textbf{78.4} & \textbf{82.4} & \textbf{82.8} & \textbf{82.0} & \textbf{70.3} & improve \\
}
\gdef\TabModulos{%
of2-bbh-q01 & tracking & Qwen 1.5B & 100.0 & 28 & 1,955 & 33 & 21 & 35.9 \\
of2-bbh-q02 & tracking & Qwen 1.5B & 100.0 & 12 & 1,722 & 42 & 22 & 41.1 \\
of2-bbh-q03 & tracking & Qwen 1.5B & 95.0 & 21 & 10,207 & 32 & 23 & 72.5 \\
of2-bbh-q04 & tracking & Qwen 1.5B & 100.0 & 16 & 3,313 & 64 & 18 & 66.3 \\
of2-bbh-q05 & tracking & Qwen 1.5B & 100.0 & 21 & 826 & 30 & 18 & 28.2 \\
of2-bbh-g01 & tracking & Gemma 1B & 100.0 & 21 & 3,499 & 29 & 17 & 46.6 \\
of2-bbh-g03 & tracking & Gemma 1B & 100.0 & 19 & 1,900 & 40 & 19 & 54.5 \\
of2-bbh-g04 & tracking & Gemma 1B & 95.4 & 23 & 7,703 & 30 & 22 & 61.6 \\
of-wol-q01 & web\_of\_lies & Qwen 1.5B & 100.0 & 21 & 31 & 4 & 1 & 2.4 \\
of-wol-q02 & web\_of\_lies & Qwen 1.5B & 100.0 & 21 & 82 & 32 & 1 & 16.1 \\
of-wol-q03 & web\_of\_lies & Qwen 1.5B & 100.0 & 19 & 18 & 6 & 1 & 4.9 \\
of-wol-q04 & web\_of\_lies & Qwen 1.5B & 100.0 & 23 & 35 & 8 & 2 & 6.5 \\
of-wol-q05 & web\_of\_lies & Qwen 1.5B & 100.0 & 17 & 31 & 1 & 1 & 1.5 \\
of-wol-g01 & web\_of\_lies & Gemma 1B & 100.0 & 21 & 32 & 40 & 3 & 35.7 \\
of-wol-g02 & web\_of\_lies & Gemma 1B & 100.0 & 18 & 18 & 69 & 3 & 82.8 \\
of-wol-g03 & web\_of\_lies & Gemma 1B & 100.0 & 19 & 18 & 37 & 3 & 46.5 \\
of-wol-g04 & web\_of\_lies & Gemma 1B & 100.0 & 13 & 72 & 1 & 3 & 1.9 \\
of-wol-g05 & web\_of\_lies & Gemma 1B & 100.0 & 17 & 19 & 3 & 3 & 3.1 \\
of-prod-q01 & product & Qwen 1.5B & 100.0 & 8 & 6,252 & 100 & 12 & 76.2 \\
of-prod-q02 & product & Qwen 1.5B & 100.0 & 20 & 6,696 & 239 & 12 & 146.4 \\
of-prod-q03 & product & Qwen 1.5B & 100.0 & 9 & 19,589 & 232 & 11 & 197.3 \\
of-prod-q04 & product & Qwen 1.5B & 100.0 & 8 & 6,395 & 84 & 13 & 69.7 \\
of-prod-q05 & product & Qwen 1.5B & 100.0 & 7 & 15,554 & 64 & 10 & 103.6 \\
of-prod-g01 & product & Gemma 1B & 100.0 & 11 & 18,284 & 56 & 14 & 106.9 \\
of-prod-g02 & product & Gemma 1B & 100.0 & 13 & 6,890 & 55 & 14 & 65.9 \\
of-prod-g03 & product & Gemma 1B & 100.0 & 13 & 8,010 & 52 & 14 & 67.9 \\
of-prod-g04 & product & Gemma 1B & 100.0 & 14 & 6,694 & 52 & 13 & 63.3 \\
of-prod-g05 & product & Gemma 1B & 100.0 & 10 & 20,870 & 54 & 9 & 113.6 \\
of-nav-q01 & navigate & Qwen 1.5B & 100.0 & 22 & 332 & 11 & 7 & 12.6 \\
of-nav-q02 & navigate & Qwen 1.5B & 100.0 & 27 & 804 & 29 & 7 & 24.5 \\
of-nav-q03 & navigate & Qwen 1.5B & 100.0 & 13 & 3,029 & 8 & 8 & 37.8 \\
of-nav-q05 & navigate & Qwen 1.5B & 100.0 & 17 & 598 & 4 & 7 & 11.6 \\
of-nav-g01 & navigate & Gemma 1B & 100.0 & 31 & 230 & 23 & 7 & 29.3 \\
of-nav-g02 & navigate & Gemma 1B & 100.0$^\dagger$ & 23 & 507 & 7 & 7 & 16.9 \\
of-nav-g03 & navigate & Gemma 1B & 100.0$^\dagger$ & 18 & 772 & 7 & 9 & 11.4 \\
of-nav-g04 & navigate & Gemma 1B & 100.0 & 14 & 1,350 & 23 & 18 & 24.8 \\
of-nav-g05 & navigate & Gemma 1B & 100.0 & 34 & 928 & 7 & 8 & 11.6 \\
of-bool-q02 & boolean\_expr & Qwen 1.5B & 100.0 & 25 & 10,896 & 343 & 26 & 338.1 \\
of-bool-q03 & boolean\_expr & Qwen 1.5B & 100.0 & 17 & 2,613 & 100 & 25 & 78.9 \\
of-bool-q05 & boolean\_expr & Qwen 1.5B & 100.0 & 22 & 1,508 & 141 & 22 & 118.0 \\
of-bool-g01 & boolean\_expr & Gemma 1B & 100.0 & 15 & 1,325 & 31 & 24 & 40.8 \\
of-bool-g02 & boolean\_expr & Gemma 1B & 100.0$^\dagger$ & 17 & 2,662 & 43 & 25 & 60.2 \\
of-bool-g03 & boolean\_expr & Gemma 1B & 100.0$^\dagger$ & 25 & 1,442 & 161 & 23 & 196.5 \\
of-bool-g04 & boolean\_expr & Gemma 1B & 100.0 & 19 & 447 & 39 & 26 & 50.2 \\
of-bool-g05 & boolean\_expr & Gemma 1B & 100.0 & 26 & 2,897 & 29 & 22 & 43.8 \\
}
\gdef\TabRondaUno{%
of-bbh-q01 & tracking, r.1 & Qwen 1.5B & 100.0 & 16 & 561 & 26 & 18 & 27.9 \\
of-bbh-q02 & tracking, r.1 & Qwen 1.5B & 100.0 & 13 & 963 & 31 & 22 & 30.2 \\
of-bbh-q03 & tracking, r.1 & Qwen 1.5B & 100.0 & 19 & 737 & 46 & 20 & 36.8 \\
of-bbh-q04 & tracking, r.1 & Qwen 1.5B & 100.0 & 18 & 1,191 & 47 & 21 & 39.5 \\
of-bbh-q05 & tracking, r.1 & Qwen 1.5B & -- & 9 & 606 & 35 & 21 & 30.5 \\
of-bbh-g01 & tracking, r.1 & Gemma 1B & 100.0 & 21 & 1,056 & 44 & 20 & 51.2 \\
of-bbh-g02 & tracking, r.1 & Gemma 1B & 100.0 & 18 & 922 & 36 & 17 & 45.8 \\
of-bbh-g03 & tracking, r.1 & Gemma 1B & 100.0 & 16 & 145 & 38 & 18 & 45.8 \\
of-bbh-g04 & tracking, r.1 & Gemma 1B & 100.0 & 15 & 994 & 24 & 14 & 31.4 \\
of-bbh-g05 & tracking, r.1 & Gemma 1B & 100.0 & 22 & 176 & 32 & 17 & 37.9 \\
}
\gdef\TabCurva{%
addition (Qwen 1.5B) & 3.0 (2/4) & v2 & 2.0 (5/5) \\
addition (Qwen 0.5B) & 3.0 (1/1) & v1 & 3.0 (5/5) \\
product (Qwen 1.5B) & 4.5 (2/2) & v3 & 3.0 (5/5) \\
product (Gemma 1B) & 3.0 (1/1) & v3.2 & 2.0 (5/5) \\
tracking (Qwen 1.5B) & 6.0 (5/5) & v3.3 & 7.0 (3/5) \\
tracking (Gemma 1B) & 10.5 (4/4) & v3.3 & 11.5 (2/5) \\
web\_of\_lies (Qwen 1.5B) & 4.0 (2/2) & v3.1 & 3.0 (5/5) \\
web\_of\_lies (Gemma 1B) & 21.0 (1/1) & v3.3 & 10.0 (5/5) \\
}
\gdef\TabTestDos{%
tracking & Qwen 1.5B & 99.1/99.4/99.2 & 99.3/99.6/99.7 & 95.7/93.7/92.7 & 100.0/86.4/80.9 \\
tracking & Gemma 1B & 99.8/99.8/99.2 (3/5) & 99.6/99.5/99.3 & 99.9/99.8/99.7 & 100.0/24.8/15.6 \\
web\_of\_lies & Qwen 1.5B & 100.0 & 100.0 & 100.0 & 99.9 \\
web\_of\_lies & Gemma 1B & 100.0 & 100.0 & 99.4 & 100.0 \\
navigate & Qwen 1.5B & 100.0 (4/5) & 100.0 & 100.0 & 95.1 \\
navigate & Gemma 1B & 100.0 & 100.0 & 91.7 & 92.3 \\
boolean\_expr & Qwen 1.5B & 100.0 (3/5) & 100.0 & 100.0 & 99.3 \\
boolean\_expr & Gemma 1B & 100.0 & 100.0 & 100.0 & 98.3 \\
product, absent lengths & Qwen 1.5B & 100.0 & -- & -- & 1.3 \\
product, absent lengths & Gemma 1B & 100.0 & -- & -- & 2.1 \\
product, 13--20 digits & Qwen 1.5B & 100.0 & -- & -- & 0.5 \\
product, 13--20 digits & Gemma 1B & 100.0 & -- & -- & 0.4 \\
}
\gdef\TabTestCC{%
composite & Qwen 1.5B & 41.05 & +21.10 & $0$ & no significant change (23 tests) \\
composite & Gemma 1B & 32.86 & +22.58 & $0$ & no significant change (23 tests) \\
nested, 3 tasks & Qwen 1.5B & 36.08 & -- & $0$ & no H10 (E2-X8) \\
nested, 3 tasks & Gemma 1B & 26.76 & -- & $0$ & no H10 (E2-X8) \\
}
\gdef\TabHDiez{%
\texttt{cc-qwen15} & composite & Qwen 1.5B & $0$ & $0$ & 0.00 & no significant change (23 tests) & success (i--iii) \\
\texttt{of-bool-q02} & boolean\_expr & Qwen 1.5B & $0$ & $0$ & 0.00 & no significant change (23 tests) & success (i--iii) \\
\texttt{of-bool-q03} & boolean\_expr & Qwen 1.5B & $0$ & $0$ & 0.00 & no significant change (23 tests) & success (i--iii) \\
\texttt{of-bool-q05} & boolean\_expr & Qwen 1.5B & $0$ & $0$ & 0.00 & no significant change (23 tests) & success (i--iii) \\
\texttt{of-bool-q01} & boolean\_expr & Qwen 1.5B & -- & -- & -- & not run (failed session) & not run \\
\texttt{of-bool-q04} & boolean\_expr & Qwen 1.5B & -- & -- & -- & not run (failed session) & not run \\
\texttt{of-nav-q01} & navigate & Qwen 1.5B & $0$ & $0$ & 0.00 & no significant change (23 tests) & success (i--iii) \\
\texttt{of-nav-q02} & navigate & Qwen 1.5B & $3.6\times10^{-4}$ & $1.6\times10^{-2}$ & 2.61 & 1 of 23 tests fail (iii) & failure \\
\texttt{of-nav-q03} & navigate & Qwen 1.5B & $0$ & $0$ & 0.00 & no significant change (23 tests) & success (i--iii) \\
\texttt{of-nav-q05} & navigate & Qwen 1.5B & $0$ & $0$ & 0.00 & no significant change (23 tests) & success (i--iii) \\
\texttt{of-nav-q04} & navigate & Qwen 1.5B & -- & -- & -- & not run (failed session) & not run \\
\texttt{of-prod-q01} & product & Qwen 1.5B & $0$ & $0$ & 0.00 & no significant change (23 tests) & success (i--iii) \\
\texttt{of-prod-q02} & product & Qwen 1.5B & $0$ & $0$ & 0.00 & no significant change (23 tests) & success (i--iii) \\
\texttt{of-prod-q03} & product & Qwen 1.5B & $0$ & $0$ & 0.00 & no significant change (23 tests) & success (i--iii) \\
\texttt{of-prod-q04} & product & Qwen 1.5B & $0$ & $0$ & 0.00 & no significant change (23 tests) & success (i--iii) \\
\texttt{of-prod-q05} & product & Qwen 1.5B & $0$ & $0$ & 0.00 & no significant change (23 tests) & success (i--iii) \\
\texttt{of2-bbh-q01} & tracking & Qwen 1.5B & $0$ & $0$ & 0.00 & no significant change (23 tests) & success (i--iii) \\
\texttt{of2-bbh-q02} & tracking & Qwen 1.5B & $0$ & $0$ & 0.00 & no significant change (23 tests) & success (i--iii) \\
\texttt{of2-bbh-q03} & tracking & Qwen 1.5B & $0$ & $0$ & 0.00 & no significant change (23 tests) & success (i--iii) \\
\texttt{of2-bbh-q04} & tracking & Qwen 1.5B & $0$ & $0$ & 0.00 & no significant change (23 tests) & success (i--iii) \\
\texttt{of2-bbh-q05} & tracking & Qwen 1.5B & $0$ & $0$ & 0.00 & no significant change (23 tests) & success (i--iii) \\
\texttt{of-wol-q01} & web\_of\_lies & Qwen 1.5B & $0$ & $0$ & 0.00 & no significant change (23 tests) & success (i--iii) \\
\texttt{of-wol-q02} & web\_of\_lies & Qwen 1.5B & $0$ & $0$ & 0.00 & no significant change (23 tests) & success (i--iii) \\
\texttt{of-wol-q03} & web\_of\_lies & Qwen 1.5B & $0$ & $0$ & 0.00 & no significant change (23 tests) & success (i--iii) \\
\texttt{of-wol-q04} & web\_of\_lies & Qwen 1.5B & $8.2\times10^{-12}$ & $0$ & 0.00 & no significant change (23 tests) & success (i--iii) \\
\texttt{of-wol-q05} & web\_of\_lies & Qwen 1.5B & $2.4\times10^{-9}$ & $1.9\times10^{-6}$ & 0.13 & not measured: changes an original weight (E2-X11) & panel not measured \\
\texttt{cc-gemma1} & composite & Gemma 1B & $0$ & $0$ & 0.00 & no significant change (23 tests) & success (i--iii) \\
\texttt{of-bool-g01} & boolean\_expr & Gemma 1B & $0$ & $0$ & 0.00 & no significant change (23 tests) & success (i--iii) \\
\texttt{of-bool-g02} & boolean\_expr & Gemma 1B & $0$ & $0$ & 0.00 & no significant change (23 tests) & success (i--iii) \\
\texttt{of-bool-g03} & boolean\_expr & Gemma 1B & $0$ & $0$ & 0.00 & no significant change (23 tests) & success (i--iii) \\
\texttt{of-bool-g04} & boolean\_expr & Gemma 1B & $0$ & $0$ & 0.00 & no significant change (23 tests) & success (i--iii) \\
\texttt{of-bool-g05} & boolean\_expr & Gemma 1B & $0$ & $0$ & 0.00 & no significant change (23 tests) & success (i--iii) \\
\texttt{of-nav-g01} & navigate & Gemma 1B & $4.2\times10^{-4}$ & $4.9\times10^{-2}$ & 44.31 & no significant change (23 tests) & failure \\
\texttt{of-nav-g02} & navigate & Gemma 1B & $0$ & $0$ & 0.00 & no significant change (23 tests) & success (i--iii) \\
\texttt{of-nav-g03} & navigate & Gemma 1B & $0$ & $0$ & 0.00 & no significant change (23 tests) & success (i--iii) \\
\texttt{of-nav-g04} & navigate & Gemma 1B & $0$ & $0$ & 0.00 & no significant change (23 tests) & success (i--iii) \\
\texttt{of-nav-g05} & navigate & Gemma 1B & $0$ & $0$ & 0.00 & no significant change (23 tests) & success (i--iii) \\
\texttt{of-prod-g01} & product & Gemma 1B & $0$ & $0$ & 0.00 & 1 of 23 tests fail (iii) & failure \\
\texttt{of-prod-g02} & product & Gemma 1B & $-2.0\times10^{-10}$ & $1.9\times10^{-6}$ & 0.05 & 1 of 23 tests fail (iii) & failure \\
\texttt{of-prod-g03} & product & Gemma 1B & $0$ & $0$ & 0.00 & no significant change (23 tests) & success (i--iii) \\
\texttt{of-prod-g04} & product & Gemma 1B & $0$ & $0$ & 0.00 & no significant change (23 tests) & success (i--iii) \\
\texttt{of-prod-g05} & product & Gemma 1B & $0$ & $0$ & 0.00 & no significant change (23 tests) & success (i--iii) \\
\texttt{of2-bbh-g01} & tracking & Gemma 1B & $0$ & $0$ & 0.00 & no significant change (23 tests) & success (i--iii) \\
\texttt{of2-bbh-g03} & tracking & Gemma 1B & $0$ & $0$ & 0.00 & no significant change (23 tests) & success (i--iii) \\
\texttt{of2-bbh-g04} & tracking & Gemma 1B & $0$ & $0$ & 0.00 & no significant change (23 tests) & success (i--iii) \\
\texttt{of2-bbh-g02} & tracking & Gemma 1B & -- & -- & -- & not run (failed session) & not run \\
\texttt{of2-bbh-g05} & tracking & Gemma 1B & -- & -- & -- & not run (failed session) & not run \\
\texttt{of-wol-g01} & web\_of\_lies & Gemma 1B & $0$ & $0$ & 0.00 & no significant change (23 tests) & success (i--iii) \\
\texttt{of-wol-g02} & web\_of\_lies & Gemma 1B & $0$ & $0$ & 0.00 & no significant change (23 tests) & success (i--iii) \\
\texttt{of-wol-g03} & web\_of\_lies & Gemma 1B & $0$ & $0$ & 0.00 & no significant change (23 tests) & success (i--iii) \\
\texttt{of-wol-g04} & web\_of\_lies & Gemma 1B & $0$ & $0$ & 0.00 & no significant change (23 tests) & success (i--iii) \\
\texttt{of-wol-g05} & web\_of\_lies & Gemma 1B & $0$ & $0$ & 0.00 & no significant change (23 tests) & success (i--iii) \\
}
\gdef\TabHOnce{%
Qwen 1.5B & boolean\_expr & 100.0 (3/5) & 100.0 & 100.0 & 100.0 & 99.3 & 97.3 \\
Qwen 1.5B & navigate & 100.0 (4/5) & 100.0 & 100.0 & 100.0 & 95.1 & 76.2 \\
Qwen 1.5B & product, absent & 100.0 & -- & 100.0 & -- & 1.3 & -- \\
Qwen 1.5B & tracking 3 & 99.1 & 95.7 & 100.0 & 100.0 & 100.0 & 48.4 \\
Qwen 1.5B & tracking 5 & 99.4 & 93.7 & 100.0 & 100.0 & 86.4 & 35.0 \\
Qwen 1.5B & tracking 7 & 99.2 & 92.7 & 100.0 & 100.0 & 80.9 & 28.5 \\
Qwen 1.5B & web\_of\_lies & 100.0 & 100.0 & 100.0 & 100.0 & 99.9 & 55.8 \\
Qwen 1.5B & damage: KL & \multicolumn{2}{l}{$1.7\times10^{-5}$ (mean of modules)} & \multicolumn{2}{l}{$0$} & \multicolumn{2}{l}{$1.3\times10^{-1}$} \\
\multicolumn{8}{l}{Qwen 1.5B: B5, B7 and B8 not measured (not run in the stage-2 exam)}\\
\midrule
Gemma 1B & boolean\_expr & 100.0 & 100.0 & 100.0 & 100.0 & 98.3 & 95.8 \\
Gemma 1B & navigate & 100.0 & 91.7 & 100.0 & 78.3 & 92.3 & 77.1 \\
Gemma 1B & product, absent & 100.0 & -- & -- & -- & 2.1 & -- \\
Gemma 1B & tracking 3 & 99.8 (3/5) & 99.9 & 100.0 & 100.0 & 100.0 & 40.8 \\
Gemma 1B & tracking 5 & 99.8 (3/5) & 99.8 & 100.0 & 100.0 & 24.8 & 20.0 \\
Gemma 1B & tracking 7 & 99.2 (3/5) & 99.7 & 100.0 & 100.0 & 15.6 & 12.9 \\
Gemma 1B & web\_of\_lies & 100.0 & 99.4 & 100.0 & 97.1 & 100.0 & 53.6 \\
Gemma 1B & damage: KL & \multicolumn{2}{l}{$1.8\times10^{-5}$ (mean of modules)} & \multicolumn{2}{l}{$0$} & \multicolumn{2}{l}{$1.2\times10^{0}$} \\
\multicolumn{8}{l}{Gemma 1B: B5, B7 and B8 not measured (not run in the stage-2 exam)}\\
}
\gdef\TabHNueve{%
Qwen 1.5B & shuffled assignment = no & tracking 5 100.0; tracking 7 100.0; tracking 3 100.0 \\
Qwen 1.5B & shuffled assignment = yes & tracking 5 100.0; tracking 7 100.0; tracking 3 100.0 \\
Qwen 1.5B & shuffled sentences = no & web\_of\_lies 100.0 \\
Qwen 1.5B & shuffled sentences = yes & web\_of\_lies 100.0 \\
Qwen 1.5B & header = Answer choices: & tracking 5 100.0; tracking 7 100.0; tracking 3 100.0 \\
Qwen 1.5B & header = Options: & tracking 5 100.0; tracking 7 100.0; tracking 3 100.0 \\
Qwen 1.5B & connectives = 0 & tracking 5 100.0; tracking 7 100.0; tracking 3 100.0 \\
Qwen 1.5B & connectives = 1 & tracking 5 100.0; tracking 7 100.0; tracking 3 100.0 \\
Qwen 1.5B & scene = band & tracking 5 100.0; tracking 7 100.0; tracking 3 100.0 \\
Qwen 1.5B & scene = cafe & tracking 5 100.0; tracking 7 100.0; tracking 3 100.0 \\
Qwen 1.5B & scene = car & tracking 5 100.0; tracking 7 100.0; tracking 3 100.0 \\
Qwen 1.5B & scene = hat & tracking 5 100.0; tracking 7 100.0; tracking 3 100.0 \\
Qwen 1.5B & scene = pet & tracking 5 100.0; tracking 7 100.0; tracking 3 100.0 \\
Qwen 1.5B & instructions = 10 & navigate 100.0 \\
Qwen 1.5B & instructions = 3 & navigate 100.0 \\
Qwen 1.5B & instructions = 4 & navigate 100.0 \\
Qwen 1.5B & instructions = 5 & navigate 100.0 \\
Qwen 1.5B & instructions = 6 & navigate 100.0 \\
Qwen 1.5B & instructions = 7 & navigate 100.0 \\
Qwen 1.5B & instructions = 8 & navigate 100.0 \\
Qwen 1.5B & instructions = 9 & navigate 100.0 \\
Qwen 1.5B & length = 10 & boolean\_expr 100.0 \\
Qwen 1.5B & length = 7 & boolean\_expr 100.0 \\
Qwen 1.5B & length = 8 & boolean\_expr 100.0 \\
Qwen 1.5B & length = 9 & boolean\_expr 100.0 \\
Qwen 1.5B & new names = no & tracking 5 100.0; tracking 7 100.0; tracking 3 100.0 \\
Qwen 1.5B & new names = yes & tracking 5 100.0; tracking 7 100.0; tracking 3 100.0 \\
Qwen 1.5B & shuffled options = no & tracking 5 100.0; tracking 7 100.0; tracking 3 100.0 \\
Qwen 1.5B & shuffled options = yes & tracking 5 100.0; tracking 7 100.0; tracking 3 100.0 \\
Qwen 1.5B & attached parentheses = no & boolean\_expr 100.0 \\
Qwen 1.5B & attached parentheses = yes & boolean\_expr 100.0 \\
Qwen 1.5B & template = 0 & boolean\_expr 100.0 \\
Qwen 1.5B & template = 1 & boolean\_expr 100.0 \\
Qwen 1.5B & template = 2 & boolean\_expr 100.0 \\
Qwen 1.5B & template = 3 & boolean\_expr 100.0 \\
Qwen 1.5B & question form = 0 & navigate 100.0; web\_of\_lies 100.0 \\
Qwen 1.5B & question form = 1 & navigate 100.0; web\_of\_lies 100.0 \\
Qwen 1.5B & question form = 2 & navigate 100.0; web\_of\_lies 100.0 \\
Qwen 1.5B & type = face\_forward & navigate 100.0 \\
Qwen 1.5B & type = turns & navigate 100.0 \\
Gemma 1B & shuffled assignment = no & tracking 5 100.0; tracking 7 100.0; tracking 3 100.0 \\
Gemma 1B & shuffled assignment = yes & tracking 5 100.0; tracking 7 100.0; tracking 3 100.0 \\
Gemma 1B & shuffled sentences = no & web\_of\_lies 97.0 \\
Gemma 1B & shuffled sentences = yes & web\_of\_lies 97.2 \\
Gemma 1B & header = Answer choices: & tracking 5 100.0; tracking 7 100.0; tracking 3 100.0 \\
Gemma 1B & header = Options: & tracking 5 100.0; tracking 7 100.0; tracking 3 100.0 \\
Gemma 1B & connectives = 0 & tracking 5 100.0; tracking 7 100.0; tracking 3 100.0 \\
Gemma 1B & connectives = 1 & tracking 5 100.0; tracking 7 100.0; tracking 3 100.0 \\
Gemma 1B & scene = band & tracking 5 100.0; tracking 7 100.0; tracking 3 100.0 \\
Gemma 1B & scene = cafe & tracking 5 100.0; tracking 7 100.0; tracking 3 100.0 \\
Gemma 1B & scene = car & tracking 5 100.0; tracking 7 100.0; tracking 3 100.0 \\
Gemma 1B & scene = hat & tracking 5 100.0; tracking 7 100.0; tracking 3 100.0 \\
Gemma 1B & scene = pet & tracking 5 100.0; tracking 7 100.0; tracking 3 100.0 \\
Gemma 1B & instructions = 10 & navigate 100.0 \\
Gemma 1B & instructions = 3 & navigate 90.8 \\
Gemma 1B & instructions = 4 & navigate 82.6 \\
Gemma 1B & instructions = 5 & navigate 79.2 \\
Gemma 1B & instructions = 6 & navigate 70.5 \\
Gemma 1B & instructions = 7 & navigate 71.6 \\
Gemma 1B & instructions = 8 & navigate 73.0 \\
Gemma 1B & instructions = 9 & navigate 72.0 \\
Gemma 1B & length = 10 & boolean\_expr 100.0 \\
Gemma 1B & length = 7 & boolean\_expr 100.0 \\
Gemma 1B & length = 8 & boolean\_expr 100.0 \\
Gemma 1B & length = 9 & boolean\_expr 100.0 \\
Gemma 1B & new names = no & tracking 5 100.0; tracking 7 100.0; tracking 3 100.0 \\
Gemma 1B & new names = yes & tracking 5 100.0; tracking 7 100.0; tracking 3 100.0 \\
Gemma 1B & shuffled options = no & tracking 5 100.0; tracking 7 100.0; tracking 3 100.0 \\
Gemma 1B & shuffled options = yes & tracking 5 100.0; tracking 7 100.0; tracking 3 100.0 \\
Gemma 1B & attached parentheses = no & boolean\_expr 100.0 \\
Gemma 1B & attached parentheses = yes & boolean\_expr 100.0 \\
Gemma 1B & template = 0 & boolean\_expr 100.0 \\
Gemma 1B & template = 1 & boolean\_expr 100.0 \\
Gemma 1B & template = 2 & boolean\_expr 100.0 \\
Gemma 1B & template = 3 & boolean\_expr 100.0 \\
Gemma 1B & question form = 0 & navigate 78.5; web\_of\_lies 97.6 \\
Gemma 1B & question form = 1 & navigate 76.8; web\_of\_lies 96.8 \\
Gemma 1B & question form = 2 & navigate 79.6; web\_of\_lies 96.9 \\
Gemma 1B & type = face\_forward & navigate 100.0 \\
Gemma 1B & type = turns & navigate 57.7 \\
}

\newcommand{\TITLE}{Don't Train It, Write It: An AI Agent Writes New Capabilities into Language Model Weights}

\title{\vspace{-1.5cm}\TITLE}
\author{Santiago Rivadeneira Quintero\\EPFL}
\date{\vspace{-1.4em}}

\setlist{itemsep=2pt,topsep=3pt}
\begin{document}
\maketitle

\def\FigUnoCifrasA{20}
\def\FigUnoCifrasP{40}
\def\FigUnoCifrasBase{24}
\def\FigUnoTope{2,048}
\def\FigUnoCotDtipo{bfloat16}
\def\FigUnoImplDtipo{float32}
\def\FigUnoImplantes{all 5}
\def\FigUnoPractica{6}
\def\FigUnoPorLong{50}
\def\FigUnoSesiones{5}
\def\FigUnoKLTokens{1,023,000}
\def\FigUnoKLSesion{of-prod-q01}
\def\FigUnoPresupuesto{50}
\def\FigUnoPasadas{8}
\def\FigUnoCabezas{100}
\def\FigUnoNeuronas{6,252}
\def\FigUnoCapaMax{11}
\def\FigUnoCapasHost{28}
\def\FigUnoSemilla{20270122}
\def\FigUnoMacFallan{3}
\def\FigUnoMacN{5}

\vspace{-2.2em}
\begin{center}
\includegraphics[width=\textwidth]{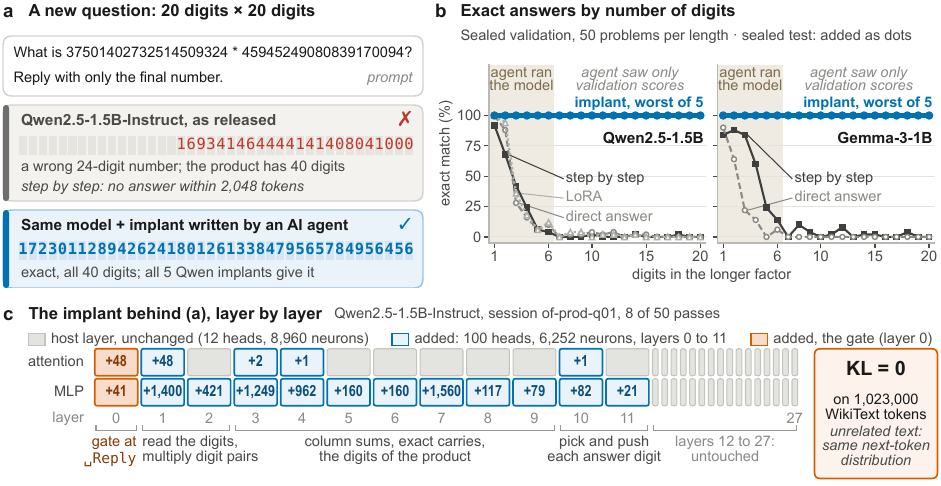}
\captionof{figure}{\textbf{An AI agent wrote \ProdLongMax-digit multiplication into Qwen2.5-1.5B-Instruct and Gemma-3-1B-it as
new weights, without training.} \textbf{(a)} A fresh \FigUnoCifrasA{}\,$\times$\,\FigUnoCifrasA-digit problem (first draw of
seed \FigUnoSemilla, in no evaluation set): the released model returns a wrong \FigUnoCifrasBase-digit number and, asked to
reason step by step (in \FigUnoCotDtipo), gives no answer within \FigUnoTope{} tokens; \FigUnoImplantes{} official Qwen
implants return all \FigUnoCifrasP{} digits (\FigUnoImplDtipo, L4 GPU, evaluation software; other backends: Section
\ref{sec:limits}). \textbf{(b)} Exact match on sealed validation by digits in the longer factor (\FigUnoPorLong{} problems per
length). The agent could run the model only in the shaded band and saw longer lengths only as one validation score per length.
Blue line: worst of the \FigUnoSesiones{} official sessions per model; blue dots: the worst examined session on the sealed
test; grey: the released model answering directly and step by step, and a LoRA trained on the practice set with the same budget
(Qwen only). \textbf{(c)} The implant of session
\texttt{\FigUnoKLSesion} (\FigUnoPasadas{} of \FigUnoPresupuesto{} passes), which answers (a): \FigUnoCabezas{} attention heads
and \FigUnoNeuronas{} MLP neurons added inside layers 0 to \FigUnoCapaMax{} of \FigUnoCapasHost; no original weight changes. On
unrelated text (\FigUnoKLTokens{} WikiText tokens) its next-token distribution equals that of the unmodified model at the same precision (KL $=0$).
\ExamenFigUno}
\label{fig:one}
\end{center}

\begin{abstract}
We show that an AI agent can write new capabilities into a pretrained language model by adding gated attention heads and MLP
neurons inside its existing layers, without backpropagation and without fitting any weight
to the answers of task examples. It has a budget of \Presupuesto{} model passes and sees no validation or test
instance. Composed in one model, its modules for six BIG-Bench Hard (BBH) subtasks raise the normalized BBH score of
Qwen2.5-1.5B-Instruct from \CCCeroQwen{} to \CCBBHQwen{} (Qwen2.5-7B-Instruct: at most \BBHSieteComun) and that
of Gemma-3-1B-it from \CCCeroGemma{} to \CCBBHGemma. Against the unmodified model at the same precision, both composites leave
the next-token distribution on \CCKLTokens{} WikiText tokens unchanged (KL exactly 0), give identical answers on
\PanelIdenN{} items of five other Leaderboard tests and identical scores on the other \PanelBBHSinModN{} BBH subtasks. The
agent also writes exact multiplication up to \ProdLongMax{} digits, running the model only up to \ProdPracticaMax{}
(\ProdTestAmbos\% on the absent lengths of the sealed test in both models; a LoRA with the same budget: at most
\BUnoProdAusMax\%). Sets, rules and report sentences were sealed in git before the first session (manifests timestamped in
Bitcoin), with dated amendments listed in Appendix D; a frozen evaluator ran the sealed test once. Verdicts (H5: tracking; H6: composite; H8: one pass vs.\ chain of
thought; H9: fresh and rewritten instances; H10: side effects; H12: multiplication), \ExamenAbstract. Not
everything worked: \HDiezModFracaso{} of \HDiezModN{} single modules show small side effects and \DosFallidas{} of
\DosPlanificadas{} sessions hit the 12-hour limit without a module.
\end{abstract}

\section{Introduction}
\label{sec:intro}

\begin{figure}[t]
\centering
\includegraphics[width=0.92\textwidth]{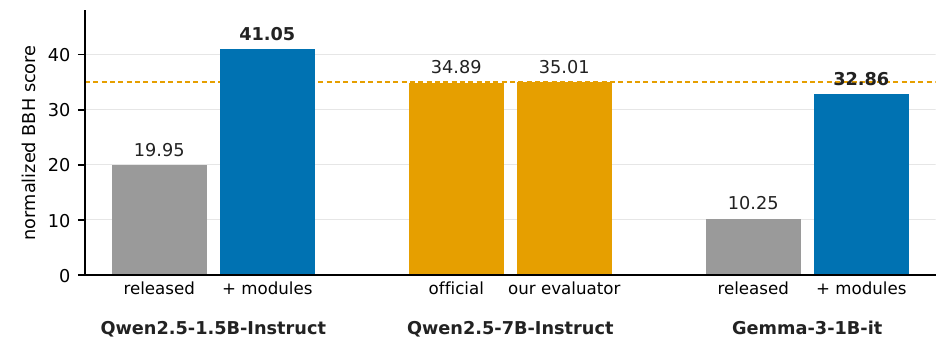}
\caption{\textbf{Agent-written modules lift a 1.5B model above its 7B sibling on BIG-Bench Hard.} Normalized BBH score (24
subtasks, protocol of the Open LLM Leaderboard v2) of each model as released (C0) and with the composite of its agent-written
modules for six subtasks, examined once on the sealed exam; the 7B reference is the official Leaderboard score and the one
from our evaluator (dashed line; \BBHSieteLmeval{} with lm-eval-harness). On the other \PanelBBHSinModN{} subtasks and the rest
of the panel, each composite matches its format control (Section \ref{sec:bbh}).}
\label{fig:titular}
\end{figure}

Large language models are improved by training: data, gradients and a loss. A recent line of work asks whether a frontier
model can instead read and write the weights of another network directly, much like code \citep{takehi2026sparks}. The
answer so far is mixed. Weight edits by agents can remove an obstacle that hides an existing skill, for example a failure
mode that wastes tokens, and they can teach facts and rules that the model can already use when they are written in its
prompt \citep{zavo2026teaching}. Whether an edit can give a model an ability it lacks even with help in its prompt was left
open \citep{zavo2026teaching}; narrow counterexamples have since appeared for addition
\citep{sudorian2026implants,mfaseehuddin2026wsp}.

We study this question with a controlled protocol and push it much further. We give a coding agent (Claude Code
\citep{anthropic2026claudecode}) tool access to the weights of a small open model and ask it to write a capability that the
model does not have. The agent writes the capability as new weights inside the network: attention heads and MLP neurons added
to existing layers, behind gates that are exactly zero off their trigger, so that on unrelated text (WikiText) the model
computes what the unmodified model computes at the same precision. It never backpropagates through the network and never fits any weight to the answers of
task examples. It has a
budget of \Presupuesto{} passes (calls to the model, each up to 512 prompts), runs the model only on practice inputs, may
request the accuracy of a sealed validation set per length or subtask, and never sees a validation or test instance. Sets,
decision rules and the sentences that would report each outcome were sealed before the first session. The agent works from a
frozen manual of procedures and lessons from development sessions. For multiplication, tracking and web\_of\_lies the manual
records designs that worked; for navigate and boolean\_expressions it states only what to compute, so in two of the five
stage-2 tasks the agent designed the circuit without the manual giving it.

\begin{figure}[t]
\centering
\input{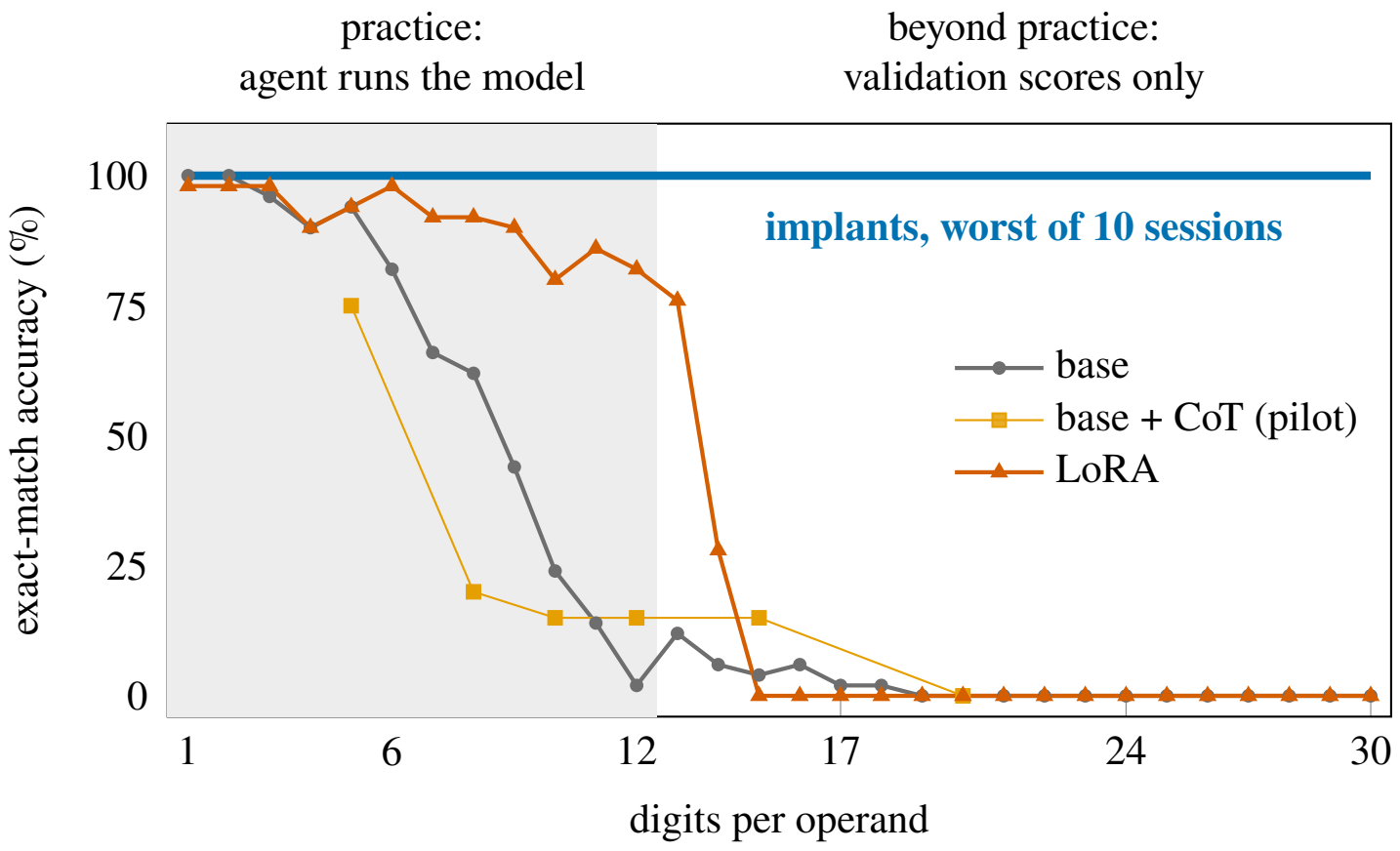}
\begin{tikzpicture}
\begin{axis}[width=0.62\textwidth, height=5.6cm, tick pos=left, xmin=0.5, xmax=30.5, ymin=-4, ymax=110, clip=false,
  xlabel={digits per operand}, ylabel={exact-match accuracy (\%)}, xtick={1,6,12,17,24,30},
  ytick={0,25,50,75,100}, tick label style={font=\scriptsize}, label style={font=\scriptsize},
  legend style={font=\scriptsize, at={(0.995,0.5)}, anchor=east, draw=none, fill=white, fill opacity=0.85, text opacity=1,
    legend columns=1, row sep=-1pt},
  legend cell align=left]
\fill[black!7] (axis cs:0.5,-4) rectangle (axis cs:\SumaPracticaMax+0.5,110);
\addplot[greycol, thick, mark=*, mark size=0.9pt] coordinates {\FigBase};
\addlegendentry{base}
\addplot[orangecol, thin, mark=square*, mark size=1.3pt] coordinates {\FigCot};
\addlegendentry{base + CoT (pilot)}
\addplot[vermcol, thick, mark=triangle*, mark size=1.2pt] coordinates {\FigLora};
\addlegendentry{LoRA}
\addplot[implcol, line width=1.6pt, mark=none, forget plot] coordinates {\FigImplante};
\node[implcol, font=\scriptsize\bfseries, anchor=north] at (axis cs:21.5,97) {implants, worst of \SumaOficiales{} sessions};
\node[font=\scriptsize, align=center, anchor=south] at (axis cs:6.5,111) {practice:\\agent runs the model};
\node[font=\scriptsize, align=center, anchor=south] at (axis cs:21.5,111) {beyond practice:\\validation scores only};
\end{axis}
\end{tikzpicture}
\caption{\textbf{Stage 1: exact addition in Qwen2.5-1.5B-Instruct.} The base model and a LoRA trained on the same practice
material lose accuracy as operands grow (sealed validation, \SumaValPorLong{} problems per length; chain of thought: a pilot
on an exploration set, \SumaCotN{} per length, capped at \SumaCotTope{} tokens); the implants of all \SumaOficiales{} official
sessions (Qwen2.5-0.5B and 1.5B) are exact at every length of the sealed test (\SumaCasosPorLong{} problems per length; the
line is the worst session), although the agent could run the model only up to \SumaPracticaMax{} digits.}
\label{fig:suma}
\end{figure}

\textbf{Results.} The headline result is a composite model. In each host, the official modules for six BBH subtasks
\citep{suzgun2023challenging} are composed, co-resident in one model without interfering, and examined once on a sealed test with the protocol of the Open LLM
Leaderboard v2 \citep{fourrier2024leaderboard}. The Qwen2.5-1.5B-Instruct composite scores \CCBBHQwen{} (normalized BBH score),
up from \CCCeroQwen; Qwen2.5-7B-Instruct scores \BBHSieteOficial{} official, \BBHSieteLmeval{} with lm-eval-harness and
\BBHSieteComun{} with our evaluator, so the 1.5B composite is above all three (Figure \ref{fig:titular}). The Gemma-3-1B-it
composite goes from \CCCeroGemma{} to \CCBBHGemma. The oracle is the gain that a perfect score on the six subtasks would
give (\OraculoQwen{} and \OraculoGemma{} points, fixed before the exam from the lm-eval baseline): the Qwen gain is close to it
and the Gemma gain reaches it. Against its \emph{format control}, the unmodified model at the same precision (float32), neither
composite changes anything else we measured: KL exactly 0 and no change of the top token on \FigUnoKLTokens{} WikiText tokens,
the same generation or choice on each of the \PanelIdenN{} items of IFEval, MATH Level 5, GPQA, MuSR and MMLU-Pro, and identical
scores on the other \PanelBBHSinModN{} BBH subtasks. The modules add specific capabilities without disturbing the rest; they do not make the model better in general. Module
by module, on fresh instances of the sealed test: on tracking\_shuffled\_objects, which Qwen2.5-1.5B cannot do even with the
algorithm in its prompt or with chain of thought (within 5 points of chance), \HCincoSesQwen{} Qwen sessions reach 90\% or more
on every absent subtask (median \HCincoMedianaQwen\%), as do \HCincoSesGemma{} in Gemma (median \HCincoMedianaGemma\%);
web\_of\_lies goes from \HOchoUnaQwen\% in one pass and \HOchoCotQwen\% with chain of thought to \TestWolQwen\% in Qwen;
navigate and boolean\_expressions, whose circuits the agent designed without a task manual, score \TestNavQwen\% and
\TestBoolQwen\% in Qwen (means of the \SesNavQwen{} and \SesBoolQwen{} sessions that produced a module; the other
\FallNavQwen{} and \FallBoolQwen{} ended at the 12-hour limit and count as 0 in the decision rules) and \TestNavGemma\% and \TestBoolGemma\% in Gemma; and exact multiplication, which Qwen2.5-1.5B cannot do even with chain of thought, scores \ProdTestQwen\% (Qwen) and
\ProdTestGemma\% (Gemma) at the lengths the agent never ran, where a LoRA trained with the same budget scores
\BUnoProdAusQwen\% and \BUnoProdAusGemma\% (Figure \ref{fig:one}). In stage 1, all \SumaOficiales{} official sessions write
exact addition that is 100\% on the sealed test at every length up to \SumaLongMax{} digits, although the agent could run
the model only up to \SumaPracticaMax{} digits (Figure \ref{fig:suma}). Not everything worked: \HDiezModFracaso{} of the
\HDiezModN{} single modules show small but measurable side effects, \DosFallidas{} of the \DosPlanificadas{} stage-2 sessions
ended at the 12-hour limit without a module, and some modules lose accuracy on rewritten wording. Section \ref{sec:bbh} reports
every preregistered verdict as it came out, with the sentence fixed for it before the exam.

\textbf{Contributions.}
\begin{itemize}
\item \textbf{A method for writing capabilities into weights} (Section \ref{sec:method}): an operating room in which new
  components are born silent inside existing layers, and an autonomous agent with a frozen, versioned manual, a budget of
  \Presupuesto{} passes and no access to any validation or test instance. The original weights stay intact in
  \AnfitrionIntactas{} of \AnfitrionSesiones{} official sessions.
\item \textbf{A 1.5B model above its 7B sibling on BBH, with no measured side effects} (Section \ref{sec:bbh}): composites of
  agent-written modules gain \CCSubeQwen{} (Qwen) and \CCSubeGemma{} (Gemma, over its format control) points of normalized BBH score; the Qwen composite,
  at \CCBBHQwen, is above every 7B reference (at most \BBHSieteComun); both match their format control on unrelated text and on the
  \PanelPruebasN{} other Leaderboard tests.
\item \textbf{Capabilities that extrapolate beyond what the agent could run} (Sections \ref{sec:suma} and \ref{sec:bbh}):
  addition up to \SumaLongMax{} digits from practice up to \SumaPracticaMax{} (\SumaPerfectas{} of \SumaOficiales{} sessions
  exact on a sealed test) and multiplication up to \ProdLongMax{} digits from practice up to \ProdPracticaMax{}
  (\ProdLejanasQwen\% from \ProdLejanasDesde{} to \ProdLejanasHasta{} digits in both models, against \BUnoProdLejQwen\% and
  \BUnoProdLejGemma\% for a LoRA with the same budget).
\item \textbf{Breadth across tasks and families} (Section \ref{sec:bbh}): modules for a BBH subtask absent even with chain of
  thought (tracking\_shuffled\_objects), for one the model only solves when reasoning (web\_of\_lies), for two whose circuits
  the agent designed with no task manual (navigate, boolean\_expressions), and for exact multiplication, in Qwen2.5 and
  Gemma 3.
\item \textbf{A protocol for trustworthy claims about agent-written weights} (Sections \ref{sec:protocol} and
  \ref{sec:costo}): sealed sets, preregistered rules and sentences, manifests timestamped in Bitcoin, a frozen evaluator run
  once on machines where no surgeon runs, an independent audit of how each weight was computed, the measured cost of each
  implant, and every failure reported.
\end{itemize}

\section{Method}
\label{sec:method}

\begin{figure*}[t]
\centering
\includegraphics[width=\textwidth]{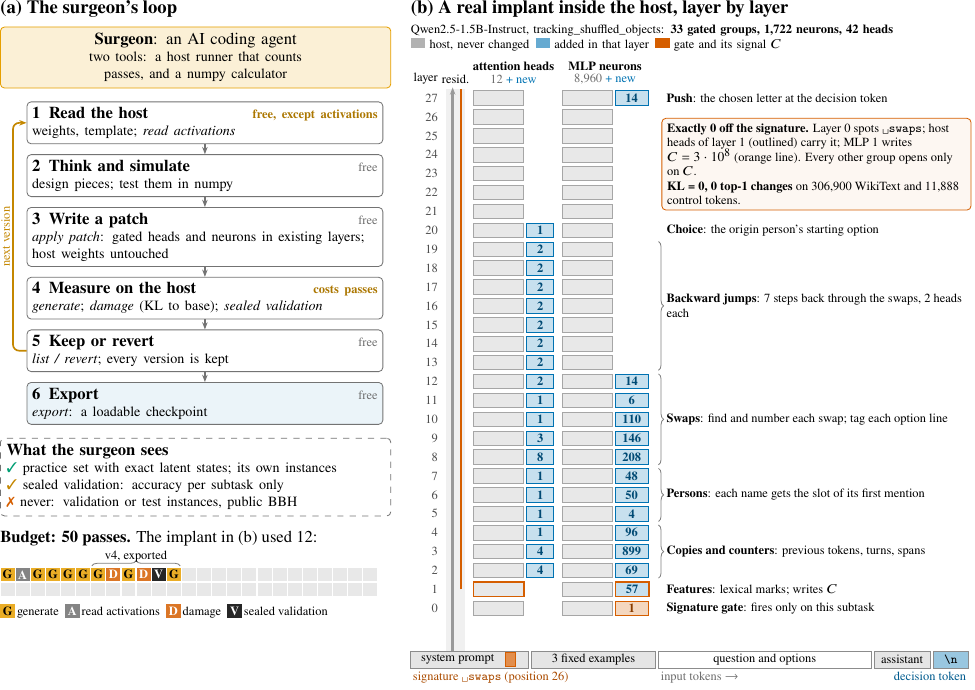}
\caption{\textbf{How the surgeon works.} \textbf{(a)} The agent reads the host's weights and activations, designs and simulates
new pieces in numpy, writes them as a patch of gated attention heads and MLP neurons inside existing layers, and measures the
patched host, which is the only step that spends its budget of \Presupuesto{} passes (one pass scores or generates up to 512
prompts, or reads activations for up to 1,024 texts); it keeps or reverts each version and exports the last one. It sees the sealed validation only as accuracy per subtask.
Tool names are translated from the original CLI. \textbf{(b)} The official tracking implant in Qwen2.5-1.5B-Instruct (session
\texttt{of2-bbh-q02}), layer by layer: host weights in grey are never changed; blue cells give the heads and neurons added in
each layer. A two-stage gate keyed to the subtask's system prompt switches every piece on; on unrelated text (WikiText)
and on control prompts each piece writes exactly 0.}
\label{fig:arq}
\end{figure*}

\subsection{The operating room}
\label{sec:quirofano}

The host model is loaded once and never trained. The agent changes it only through \emph{patches}: Python scripts that build
tensors, registered with their hash and parent so that each can be reverted. A patch may add \emph{groups} to an
existing layer: a group of new attention heads with their own query, key, value and output matrices, or a group of new MLP
neurons with their own gate, up and down projections. The output of a group is multiplied by a gate, a scalar function of
the residual stream designed by the agent, before it is added to the residual (\GruposConPuerta{} of the \GruposN{} groups of
the official implants have one). Off its trigger a gate is exactly zero by construction, so its group adds exactly zero; that
the layer then computes what the original layer computes we measure on unrelated text (WikiText), because inputs close to the
task form can still fire a trigger (Section \ref{sec:limits}). Groups that need exact arithmetic run in
float64 and round once into the float32 residual. New layers, new vocabulary and changes to the original tensors are not
needed; the protocol allows editing existing weights, but in \AnfitrionIntactas{} of \AnfitrionSesiones{} official sessions the
final implant touches none of them (the exception, \AnfitrionSesion{}, changes one matrix, \texttt{\AnfitrionTensor}, by a
Frobenius norm of \AnfitrionNorma). Figure \ref{fig:arq} shows the loop and one real implant, layer by layer.

The final implant is exported as one checkpoint: the host's configuration and tensors, the new tensors and a model file, loaded
with \texttt{trust\_remote\_code} in MLX. The model file is the operating room's layer code and is the same for every session of
a stage (SHA-256 prefix \texttt{\CodigoShaUno} in the \SumaOficiales{} addition sessions, \texttt{\CodigoShaDos} in all stage-2
sessions, across five tasks and two families). Its forward pass contains no task-specific code: it runs each host block
unchanged and adds each group's neurons or heads, with the host's activation, rotary codes and cache, multiplied by the group's
gate when it has one (a product of ReLU or step functions of affine maps of the residual), in float64 where a group is exact. What a module
computes is in its weights. All \CodigoModulos{} official implants have gates and load through this file.

The model is executed only by the operating-room tool. A \emph{pass} is the scoring or generation of up to 512 prompts, or a
read of activations for up to 1,024 texts; reading or analysing weights is free. Each session has \Presupuesto{} passes and the
tool refuses any pass beyond them. The agent sees a practice set and, for BBH, the three fixed exemplars of the evaluation
prompt. It may request the sealed-validation score, which costs one pass and returns only accuracy per length (per subtask for
BBH), never an instance; the tool refuses operands longer than the practice lengths (\SumaPracticaMax{} digits for addition,
\ProdPracticaMax{} for multiplication), so the agent never runs the model beyond them. It never sees a validation instance, a
test instance, the rewritten sets or the public BBH items: the operating room serves only practice data and the agent's file
permissions exclude the sealed files.

\subsection{The surgeon}
\label{sec:cirujano}

The surgeon is an autonomous Claude Code session \citep{anthropic2026claudecode}. In stage 2 each session runs headless on a
cloud GPU (Claude Code \CirujanoVersiones{}, model \texttt{\CirujanoModelo} in every recorded transcript, a cap of
\CirujanoTopeHoras{} hours); in stage 1 the sessions ran on a laptop with the same tools. A session receives a task brief and a
\emph{manual}: procedures and lessons from development sessions (for example, how to read operands with rotary position
codes, how to build an exact gate, how to keep a module silent in chat prose). The manual is frozen per version, with the
SHA-256 of every file and a git tag (v1, v2 for addition; v3 to v3.4 for stage 2). For multiplication, tracking and
web\_of\_lies the stage-2 manual records what development sessions found: the algorithm (for web\_of\_lies, \emph{Yes} if and
only if the number of lie words is even), the template positions that served as triggers, and the failures to avoid. For these
three tasks, official sessions measure whether a fresh agent reliably turns a documented design into exact, silent weights;
discovery happened in development sessions (Appendix \ref{app:curva}). navigate and boolean\_expressions had no development
session. For them the manual (v3.4, which is v3.3 unchanged plus a task brief) states only what the module must compute (whether
the walk ends where it started; the value of the expression, with parentheses and the precedence of \emph{not}, \emph{and},
\emph{or}), the answer format and its scoring, the rules, and general lessons on gates, silence and composition written for other
subtasks; it gives no algorithm, circuit, layer or trigger position. In these two of the five stage-2 tasks the agent designed
the circuit itself. For example, the navigate module of \texttt{of-nav-g04} parses each clause into a move or a turn, keeps the
heading modulo four turns and answers \emph{Yes} if and only if the displacements sum to zero, and the module of
\texttt{of-bool-g01} evaluates the expression level by level of parentheses, with \emph{not} before \emph{and} before \emph{or}.
Between rounds a separate agent, the librarian, may change the manual only with evidence from development sessions. A typical session diagnoses how the host represents the task,
designs a circuit in numpy, checks it in simulation, writes it as a patch, measures practice accuracy and leakage on generic
text, and iterates until it is exact on practice or the budget ends.

The rule in the surgeon's manual reads (our translation from Spanish): ``no gradients and no numerical optimizers on the weights (CMA-ES, least squares
fitted to task data); class means, LDA, projections and any numpy computation to understand and design are allowed.''
Closed-form statistics of host activations (class means, linear discriminants) serve to read token-level features, such as
which digit or name sits at a position. An independent audit, with its criterion fixed in git before any session was opened,
traced every weight of the evaluated implants of \AuditN{} official sessions (\AuditDos{} of stage 2 and the \AuditUno{} of
stage 1) back to the code that computed it (Appendix \ref{app:auditoria}): class A uses closed-form linear algebra only; class B
has a small numerical search in its final weights; class C tried one and discarded it. No session backpropagates through the
network and none fits any weight to the answers of task examples. In \AuditB{} of \AuditN{} sessions, some design constants
(rotary-kernel amplitudes, code frequencies, output push vectors, marker directions) were set by a small numerical search over
rotary geometry, the vocabulary or host activations (linear programming, random search, least squares on a grid, an iterative
minimax, or (sub)gradient ascent on a margin, the last in \AuditGrad{} sessions, against the letter of the rule); none of these
searches uses the answers of task examples, and all \AuditN{} sessions count as preregistered. Among stage-2 sessions,
validation outcome is similar across classes: \AuditACien{} of \AuditAVal{} A sessions, \AuditBCien{} of \AuditBVal{} B and
\AuditCCien{} of \AuditCVal{} C reach 100\% on sealed validation. We therefore state the method as \emph{without backpropagation, and without fitting any weight to the answers of
task examples}.

\subsection{The protocol}
\label{sec:protocol}

Producing a result and believing it are separate steps. Before the first session of each stage we sealed, with SHA-256 hashes
in git: the practice, validation and test sets (fixed seeds), the hypotheses, the decision rules and the sentence that would
report each outcome. A \emph{cell} is a task in one host model; each cell has five official sessions launched with a frozen
operating room and manual, and a \emph{round} is the five official sessions of one cell under one frozen manual version.
Development sessions are labelled, reported apart and never counted. Each official implant is evaluated once by a frozen
evaluator (its files hashed); in stage 2, on machines where no surgeon runs. There, each stage-2 implant is rebuilt on a
different machine from its saved patch tensors (the final values each patch wrote, hashed in the session's patch registry),
applied to the host by the frozen operating room; \ReexpN{} of the \DosTerminadas{} finished stage-2 implants have been rebuilt,
each with the same SHA-256 as the official export written on the surgeon's machine. A script checks, without opening any
figure, that every unit of the exam, and every instance of a panel split across machines, is counted once; panel instances that
a planning fault made two workers compute were resolved by a rule fixed blind and applied by code (Appendix
\ref{app:protocolo}). The SHA-256 manifests of the stage-1 results, of the stage-2 specification and sealed sets, and of the
first draft of this preprint with the sealed sets, rules and amendments \SelloPreprintEnmiendas, made before the exam results
were gathered, are stamped in Bitcoin with OpenTimestamps (blocks \SelloUno, \SelloDos{} and \SelloPreprint); so is a last
manifest, made just before the exam was gathered, of the repository with all amendments (\SelloPrecierreEnmiendas), the
closing code and the raw exam copies of every machine, unopened (block \SelloPrecierre; the exam was gathered after that block
was mined). Amendments are
dated and hashed in git before the exam was gathered, except E2-X30, written after it and only for presentation; two were sealed after the computation they regulate had run, though before
anyone read it: E2-X20, which let the composite's KL part, killed by memory on an L4 before writing anything, run on an H200
(moved at \XVeinteMovida{} CET, amendment sealed at \XVeinteSellada), and E2-X24, the rule for panel instances computed twice,
committed together with its first application (\XVeintiCuatroSellada{} CET). Appendix \ref{app:protocolo} lists those that touch
what we report. All official sessions of the
round that counts are reported; tracking's first official round (manual v3.1, \RondaUnoN{} sessions) was reclassified as
development by a dated amendment before the exam and is listed in Appendix \ref{app:modulos}.

\section{Stage 1: exact addition}
\label{sec:suma}

\textbf{Task.} ``What is $a + b$? Reply with only the final number.'', in the chat template, greedy decoding, operands of
equal length from 1 to \SumaLongMax{} digits. Hosts: Qwen2.5-1.5B-Instruct and Qwen2.5-0.5B-Instruct \citep{qwen2024qwen2},
which tokenize numbers digit by digit. The practice set served to the agent stops at \SumaPracticaMax{} digits; the sealed test
has \SumaCasosPorLong{} problems per length (\SumaCasosTest{} per session). Lengths where the base model scores 2\% or less on
the sealed validation are \emph{absent} (rule A6); they include every length from \SumaBaseQuinceDesde{} on in the 1.5B model
and from \SumaBaseCincoDesde{} on in the 0.5B model.

\textbf{Can the agent add a capability the model lacks?} Yes. All \SumaOficiales{} official sessions, five per model, reach
100\% at every length of the sealed test (Table \ref{tab:suma}); the preregistered rule required 3 of 5 at 90\% or more on the
absent lengths. With chain of thought (a pilot on an exploration set, \SumaCotN{} problems per length, capped at \SumaCotTope{}
tokens), Qwen2.5-1.5B solves \SumaCotQuince\% at 15 digits and \SumaCotVeinte\% at 20, where \SumaCotVeinteSinResp{} of
\SumaCotN{} answers give no final answer within the cap. The session reports of \SumaVentanaN{} of the \SumaOficiales{}
implants state a finite carry window: a run of \SumaVentanaMin{} to \SumaVentanaMax{} consecutive columns whose digits sum to 9
(the length depends on the session) can make them fail. Uniform random operands rarely contain such runs, so the implants are
exact on the sealed test, not proven exact on every input.

\textbf{Does it extrapolate?} Yes. Lengths \SumaNuncaDesde{} to \SumaLongMax{} were never run by the agent; it saw them only as
validation accuracy per length. The median session requested validation \SumaValLecturasMed{} time and none more than
\SumaValLecturasMax{} times; in \SumaValLimpias{} of \SumaOficiales{} sessions every request already returned 100\% at every
length, and in \SumaValSuciaSesion{} one request showed \SumaValSuciaDetalle{} digits and the agent revised its implant. In
every session the exported implant is the last one validated, and every session is exact on the sealed test. A LoRA
\citep{hu2021lora} trained on \LoraEjemplos{} practice examples (the budget of \Presupuesto{} passes of 512 prompts, best of
\LoraConfigs{} configurations on validation up to \SumaPracticaMax{} digits, \LoraSemillas{} seed) reaches \LoraQuinceA\%,
\LoraQuinceB\% and \LoraQuinceC\% on lengths 1--12, 13--20 and 21--30 of the sealed validation in Qwen2.5-1.5B
(\LoraCincoA\%, \LoraCincoB\% and \LoraCincoC\% in 0.5B). The LoRA shows what fitting the same practice data gives in this
host, with as many examples as the surgeon's budget; it is chosen and reported on the same validation set, which favours it,
and it is not the best trained method for length generalization. Activation steering \citep{turner2023steering} and CMA-ES
\citep{hansen2016cma} with the same budget stay at the base level (\SteerSel\% against \SteerBase\% on 1--12 digits for
steering; \CmaesC\% on 21--30 digits for CMA-ES).

\textbf{How one implant works.} Session \EjSesion{} (\EjPasadas{} passes) writes \EjNeuronas{} gated neurons and \EjCabezas{}
query heads into \EjCapas{} layers and changes no prompt position: everything is computed at the \texttt{assistant} token and
at the answer positions, which read the intact prompt by relative position. First, a head gated exactly on
\texttt{assistant} checks that the token `` Reply'' sits at its fixed offset. Next, position readers measure the operand length
$n$ and write it as a rotary phase code, and readers in the following layer rotate their query by that code to fetch, for the
current answer column, the aligned digits of $a$ and $b$. Two more layers compute generate and propagate terms over a window of
columns and carry the result to each answer position; the last one writes $(a+b+c) \bmod 10$ into the logit of the next digit,
with the exact groups in float64. A correct algorithm extrapolates by construction; what the lengths beyond \SumaPracticaMax{}
test is whether the agent's reading of the host (operand positions under rotary codes) holds where it could not run the model
and saw only the sealed validation score per length.
The host supplies tokenization, embeddings and rotary attention; the algorithm is the agent's.

\textbf{Is the rest of the model unchanged?} Where measured in full, yes. For session of-15b-01 the exported model and the
base model give the same next-token distribution at all \DanoPosiciones{} positions of \KlPasajes{} WikiText-103 passages
\citep{merity2016pointer} (KL exactly 0, \DanoCambios{} changes of the top token). MMLU \citep{hendrycks2020measuring} is
\MmluExp\% for the implant and \MmluControl\% for its format control (the same export with the implant output set to zero),
against \MmluBase\% for the base model (\MmluDifPreguntas{} question of \MmluN); ARC-Challenge \citep{clark2018think} is
\ArcExp\% in all three. For of-15b-02, the mean KL is $\DanoDosKL$ with \DanoDosCambios{} top-token changes in
\DanoPosiciones{} positions, all at near-ties of the base model. The other eight sessions have only the damage checks the
agent ran itself.

\textbf{Does it help a real benchmark?} Not yet. The implant fires on its template only. Even a perfect external calculator for
sums and differences raises GSM-HARD \citep{gao2023pal} from \GsmBase\% to \GsmSumaResta\% for Qwen2.5-1.5B with chain of
thought (\GsmN{} problems; \GsmTodas\% with all four operations), below the 5-point gain we had preregistered.

\begin{table}[t]
\centering
\caption{Stage 1, the ten official sessions. All \SumaOficiales{} reach 100\% at each of the \SumaLongMax{} lengths of the
sealed test. Validation requests (each returns accuracy per length only), forward passes used of \Presupuesto{}, and what the
final implant adds, read from the patch headers: new MLP neurons, new attention heads (query heads), layers touched, and the
entries of the new tensors in millions (dense allocation, not nonzeros). No original tensor changes in any of the ten.}
\label{tab:suma}
\small
\begin{tabular}{llrrrrrr}
\toprule
Session & Host & Val. reads & Passes & Neurons & Heads & Layers & Entries (M)\\
\midrule
\TabSuma
\bottomrule
\end{tabular}
\end{table}

\section{Stage 2: BIG-Bench Hard, multiplication and composite models}
\label{sec:bbh}

\textbf{Tasks.} Six BBH subtasks \citep{suzgun2023challenging} with a fixed template (tracking\_shuffled\_objects with 3, 5
and 7 objects, web\_of\_lies, navigate, boolean\_expressions), scored as in the protocol of the Open LLM Leaderboard v2
\citep{fourrier2024leaderboard}: chat template, three exemplars as turns, multiple choice by likelihood. Two normalizations
appear. Per item, the answer is the option with the highest length-normalized log-likelihood (\texttt{acc\_norm}, the accuracy
of Tables \ref{tab:ausencia} and \ref{tab:validacion}); per benchmark, the Leaderboard's \emph{normalized score} rescales each
subtask's accuracy from chance to 100 and averages over its 24 BBH subtasks. We reproduce that protocol with lm-eval-harness
\citep{biderman2024lessons} and with our own evaluator (normalized score of Qwen2.5-1.5B-Instruct: \BBHQwenComun{} with our
evaluator, \BBHQwenOficial{} official). H6 calls a composite above its 7B sibling only if its normalized score exceeds all three
scores of Qwen2.5-7B-Instruct: \BBHSieteOficial{} official, \BBHSieteLmeval{} with lm-eval-harness and \BBHSieteComun{} with
our evaluator (amendment E2-X29, Appendix \ref{app:protocolo}). Exact multiplication: ``What is $a \times b$? Reply with only the final number.'',
practice up to \ProdPracticaMax{} digits per operand, sealed validation and test up to \ProdLongMax{}. Hosts:
Qwen2.5-1.5B-Instruct and Gemma-3-1B-it \citep{team2025gemma}. For BBH the agent sees \BBHPracticaN{} practice instances per
subtask (fewer for boolean\_expressions) from our own sealed generator; validation (\BBHValN{} per subtask), test and rewritten
sets come from the same generator with other seeds, and the \BBHPublicoN{} public items per subtask are never shown. No
practice, validation, test or rewritten instance appears in public BBH or in the Leaderboard exemplars, and an exact solver
agrees with the public answer key on \BBHSolucionadorN{} of \BBHPublicoN{} items in every subtask. Generated items share the
public templates, so our test and the public items measure fresh instances of the same template; only the rewritten sets go
beyond it. On the public \texttt{mul\_big} problems of DeepMind Mathematics \citep{saxton2019analysing} (signed decimals, a
different format), Qwen2.5-1.5B-Instruct answers \MulBigDirecto\% directly; this set has no preregistered success rule.

\textbf{What is absent.} We fixed a criterion before operating: a cell (model, subtask) is \emph{absent} if no condition passes
chance plus 5 points: direct answer on the public set and on our validation, and chain of thought with the standard exemplars,
with relaxed stopping, and with the algorithm written in the prompt; it is a \emph{compile} cell if only chain of thought passes;
otherwise it is an \emph{improve} cell (Table \ref{tab:ausencia}). Tracking is absent in Qwen2.5-1.5B for 3, 5 and 7 objects
(for 3 objects, within 5 points of chance: \TrackTresAlg\% with the algorithm in the prompt, 95\% interval \TrackTresAlgLo{}
to \TrackTresAlgHi{}) and in Gemma for 5 and 7; web\_of\_lies is a compile cell in both. Multiplication is absent in
Qwen2.5-1.5B even with chain of thought (\ProdCotQwenAusentes\% on its \ProdAusentesQwenN{} absent lengths;
\ProdCotQwenCortadas\% of all answers hit the \ProdCotTope-token cap) and a compile cell in Gemma (\ProdCotGemmaAusentes\%). If
all six subtasks reached 100\%, the normalized score would rise by \OraculoQwen{} points in Qwen and \OraculoGemma{} in Gemma:
this is the oracle against which the composed model is judged.

\begin{table}[t]
\centering
\caption{Criterion of absence, fixed before any stage-2 session. Accuracy (\%) of the base model; bold marks values above
chance plus 5 points. Direct: one forward pass, as in the Leaderboard protocol (public set, \BBHPublicoN{} items) and on our sealed
validation (\BBHValN{} items; boolean: \BBHValBoolN). CoT: greedy chain of thought with the three original BBH exemplars, standard stopping,
relaxed stopping, and with the algorithm explained in the prompt (public set).}
\label{tab:ausencia}
\small
\setlength{\tabcolsep}{4.5pt}
\begin{tabular}{@{}llrrrrrrl@{}}
\toprule
 & & & \multicolumn{4}{c}{Public set (\BBHPublicoN)} & Sealed val. & \\
\cmidrule(lr){4-7}\cmidrule(lr){8-8}
Host & Subtask & Chance+5 & Direct & CoT std & CoT relax & CoT alg. & Direct & Class\\
\midrule
\TabAusencia
\bottomrule
\end{tabular}
\end{table}

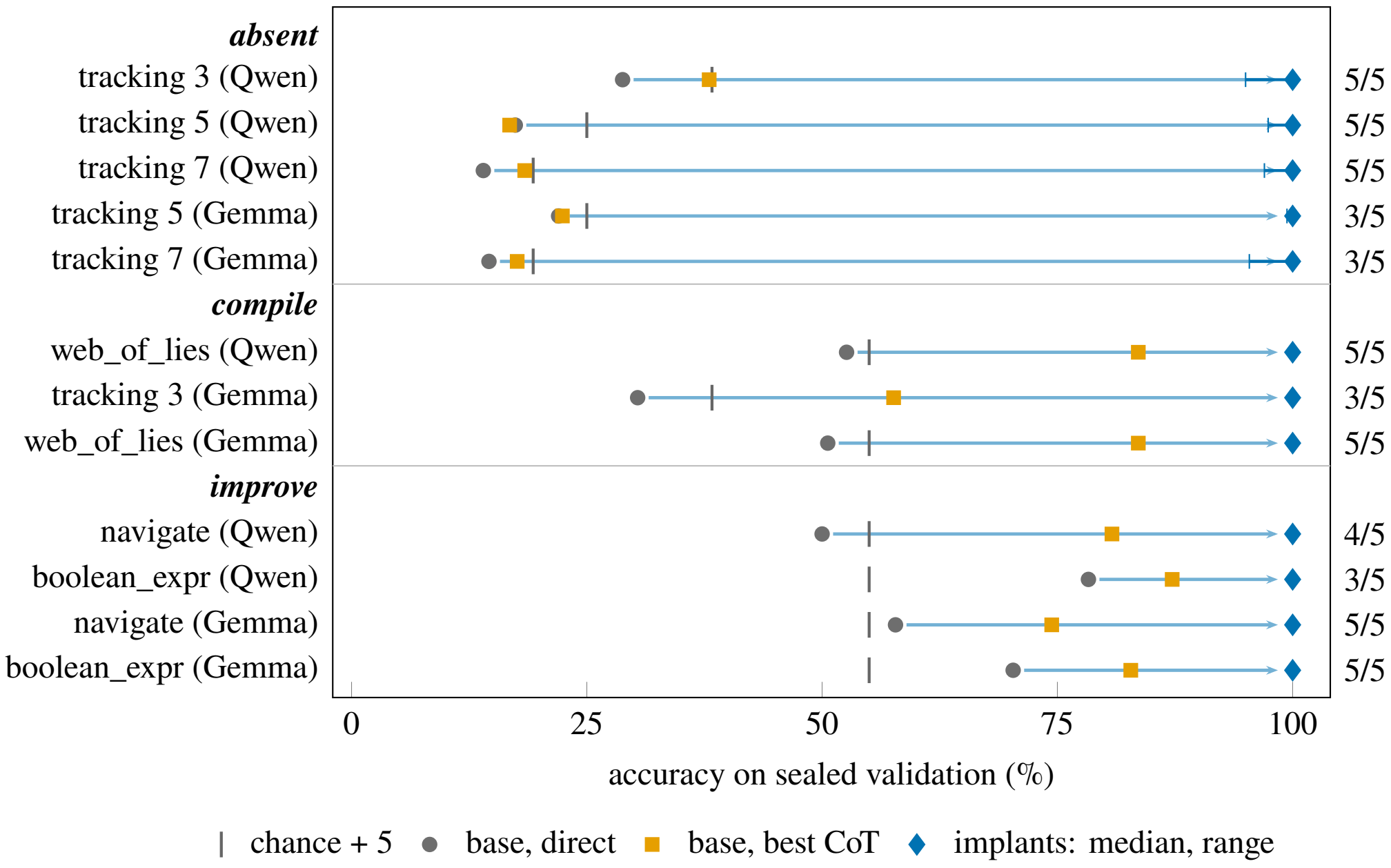
\begin{figure}[tbp]
\centering
\pgfplotsset{figdosfilas/.style={ytick={1,...,15}, yticklabels={{\textbf{\textit{absent}}},{tracking 3 (Qwen)},{tracking 5 (Qwen)},{tracking 7 (Qwen)},{tracking 5 (Gemma)},{tracking 7 (Gemma)},{\textbf{\textit{compile}}},{web\_of\_lies (Qwen)},{tracking 3 (Gemma)},{web\_of\_lies (Gemma)},{\textbf{\textit{improve}}},{navigate (Qwen)},{boolean\_expr (Qwen)},{navigate (Gemma)},{boolean\_expr (Gemma)}}, ymin=0.4, ymax=15.6}}
\def\FigDosUmbral{(38.3,2) (25.0,3) (19.3,4) (25.0,5) (19.3,6) (55.0,8) (38.3,9) (55.0,10) (55.0,12) (55.0,13) (55.0,14) (55.0,15)}
\def\FigDosBase{(28.8,2) (17.4,3) (14.0,4) (22.0,5) (14.6,6) (52.6,8) (30.4,9) (50.6,10) (50.0,12) (78.3,13) (57.8,14) (70.3,15)}
\def\FigDosCot{(38.0,2) (16.8,3) (18.4,4) (22.4,5) (17.6,6) (83.6,8) (57.6,9) (83.6,10) (80.8,12) (87.2,13) (74.4,14) (82.8,15)}
\def\FigDosImpl{(100.0,2) += (0.0,0) -= (5.0,0) (100.0,3) += (0.0,0) -= (2.6,0) (100.0,4) += (0.0,0) -= (3.0,0) (100.0,5) += (0.0,0) -= (0.6,0) (100.0,6) += (0.0,0) -= (4.6,0) (100.0,8) += (0.0,0) -= (0.0,0) (100.0,9) += (0.0,0) -= (0.2,0) (100.0,10) += (0.0,0) -= (0.0,0) (100.0,12) += (0.0,0) -= (0.0,0) (100.0,13) += (0.0,0) -= (0.0,0) (100.0,14) += (0.0,0) -= (0.0,0) (100.0,15) += (0.0,0) -= (0.0,0)}
\def\FigDosK{(105,2) [5/5] (105,3) [5/5] (105,4) [5/5] (105,5) [3/5] (105,6) [3/5] (105,8) [5/5] (105,9) [3/5] (105,10) [5/5] (105,12) [4/5] (105,13) [3/5] (105,14) [5/5] (105,15) [5/5]}
\def\FigDosClases{\draw[black!25] (axis cs:-2,6.5) -- (axis cs:104,6.5); \draw[black!25] (axis cs:-2,10.5) -- (axis cs:104,10.5);}
\def\FigDosFlechas{\draw[-{Stealth[length=3.2pt]}, implcol!55, line width=0.9pt, shorten <=3pt, shorten >=4pt] (axis cs:28.8,2) -- (axis cs:100.0,2); \draw[-{Stealth[length=3.2pt]}, implcol!55, line width=0.9pt, shorten <=3pt, shorten >=4pt] (axis cs:17.4,3) -- (axis cs:100.0,3); \draw[-{Stealth[length=3.2pt]}, implcol!55, line width=0.9pt, shorten <=3pt, shorten >=4pt] (axis cs:14.0,4) -- (axis cs:100.0,4); \draw[-{Stealth[length=3.2pt]}, implcol!55, line width=0.9pt, shorten <=3pt, shorten >=4pt] (axis cs:22.0,5) -- (axis cs:100.0,5); \draw[-{Stealth[length=3.2pt]}, implcol!55, line width=0.9pt, shorten <=3pt, shorten >=4pt] (axis cs:14.6,6) -- (axis cs:100.0,6); \draw[-{Stealth[length=3.2pt]}, implcol!55, line width=0.9pt, shorten <=3pt, shorten >=4pt] (axis cs:52.6,8) -- (axis cs:100.0,8); \draw[-{Stealth[length=3.2pt]}, implcol!55, line width=0.9pt, shorten <=3pt, shorten >=4pt] (axis cs:30.4,9) -- (axis cs:100.0,9); \draw[-{Stealth[length=3.2pt]}, implcol!55, line width=0.9pt, shorten <=3pt, shorten >=4pt] (axis cs:50.6,10) -- (axis cs:100.0,10); \draw[-{Stealth[length=3.2pt]}, implcol!55, line width=0.9pt, shorten <=3pt, shorten >=4pt] (axis cs:50.0,12) -- (axis cs:100.0,12); \draw[-{Stealth[length=3.2pt]}, implcol!55, line width=0.9pt, shorten <=3pt, shorten >=4pt] (axis cs:78.3,13) -- (axis cs:100.0,13); \draw[-{Stealth[length=3.2pt]}, implcol!55, line width=0.9pt, shorten <=3pt, shorten >=4pt] (axis cs:57.8,14) -- (axis cs:100.0,14); \draw[-{Stealth[length=3.2pt]}, implcol!55, line width=0.9pt, shorten <=3pt, shorten >=4pt] (axis cs:70.3,15) -- (axis cs:100.0,15);}
\begin{tikzpicture}
\begin{axis}[figdosfilas, y dir=reverse, width=0.70\textwidth, height=8.2cm, xmin=-2, xmax=104, clip=false, tick pos=left,
  xlabel={accuracy on sealed validation (\%)}, xtick={0,25,50,75,100}, ytick style={draw=none},
  tick label style={font=\footnotesize}, label style={font=\footnotesize}, yticklabel style={font=\footnotesize},
  legend style={font=\footnotesize, at={(0.42,-0.17)}, anchor=north, draw=none, legend columns=4, column sep=6pt},
  legend cell align=left]
\FigDosClases
\FigDosFlechas
\addplot[only marks, mark=|, mark size=3.5pt, black!60, thick] coordinates {\FigDosUmbral};
\addlegendentry{chance + 5}
\addplot[only marks, mark=*, greycol, mark size=1.9pt] coordinates {\FigDosBase};
\addlegendentry{base, direct}
\addplot[only marks, mark=square*, orangecol, mark size=1.8pt] coordinates {\FigDosCot};
\addlegendentry{base, best CoT}
\addplot[only marks, mark=diamond*, implcol, mark size=2.8pt, error bars/.cd, x dir=both, x explicit,
  error bar style={implcol, thick}, error mark options={rotate=90, implcol, mark size=2pt}] coordinates {\FigDosImpl};
\addlegendentry{implants: median, range}
\addplot[only marks, mark=none, nodes near coords, point meta=explicit symbolic,
  nodes near coords style={anchor=west, font=\footnotesize, inner sep=1pt}, forget plot] coordinates {\FigDosK};
\end{axis}
\end{tikzpicture}
\caption{\textbf{Stage 2 on BIG-Bench Hard, sealed validation.} Each row is a BBH subtask in one host, grouped by the class of
Table \ref{tab:ausencia}. The arrow goes from the base model answering directly on sealed validation (grey) to the median of
the implants of the cell's finished official sessions on that subtask (blue: median, bar from worst to best; $k/5$: sessions
with a score); orange is the base model's best chain of thought on the public set (best of the three prompts of Table
\ref{tab:ausencia}), and the tick marks chance plus 5 points, the threshold of absence. Implant values are sealed validation,
which the agent could request as accuracy per subtask (Section \ref{sec:quirofano}); \EstadoFigDos. Exact multiplication, the third stage-2 task, is in Figure \ref{fig:one}b.}
\label{fig:dos}
\end{figure}

\textbf{Results on sealed validation.} Of \DosPlanificadas{} planned official sessions, \DosTerminadas{} have finished with
an exported module, \DosFallidas{} failed (12 h limit, no export; each counts in its five as a session that reaches no
threshold), \DosValMedidas{} have a validation score and \DosValCien{} of those reach 100\% (Table \ref{tab:validacion}; Figure
\ref{fig:dos} for BBH and Figure \ref{fig:one}b for multiplication). Tracking modules reach 100\% on all three lengths in \ValCienTrackQwen{} of \TermTrackQwen{} finished sessions
in Qwen and \ValCienTrackGemma{} of \TermTrackGemma{} in Gemma, whose other \FallTrackGemma{} failed; product modules are exact from 1 to \ProdLongMax{} digits in
\ValCienProdQwen{} of \TermProdQwen{} sessions in Qwen and \ValCienProdGemma{} of \TermProdGemma{} in Gemma, although the agent
could run the model only up to \ProdPracticaMax{} digits and saw longer lengths only as one sealed validation score per length; web\_of\_lies modules are exact in \ValCienWolQwen{} of
\TermWolQwen{} and \ValCienWolGemma{} of \TermWolGemma. navigate and boolean\_expressions modules, whose circuits the agent
designed without a task manual (Section \ref{sec:cirujano}), are exact in \ValCienNavQwen{} of \TermNavQwen{} and
\ValCienNavGemma{} of \TermNavGemma{} navigate sessions and in \ValCienBoolQwen{} of \TermBoolQwen{} and \ValCienBoolGemma{} of
\TermBoolGemma{} boolean\_expressions sessions, in Qwen and Gemma respectively (\SinManualXCuatro{} of them scored by us on the
exported model, because an infrastructure fault denied the session's request). In \ProdValLimpias{} of \ProdValN{} product sessions every validation
request returned 100\% at every length; \ProdValSuciaSesion{} saw \ProdValSuciaMin{} to \ProdValSuciaMax\% between
\ProdValSuciaDesde{} and \ProdValSuciaHasta{} digits in one request and revised its implant. In all \DosValUltima{} stage-2
sessions that requested validation, the exported implant is the last one validated (at most \DosValLecturasMax{} requests per
session).
These numbers come from a sealed set the agent never saw item by item, but it could request the aggregate score and stop when
it was high: they are an optimistically selected estimate, not a substitute for the test, which follows. In the documented
tasks, what the
modules compute follows the manual: web\_of\_lies modules count lie words and answer by their parity, tracking modules recover the final assignment from the
sequence of swaps, and BBH modules open their gates on the fixed system message and exemplars of their subtask, which the
evaluation prompt repeats for every instance.

\begin{table}[t]
\centering
\caption{Stage 2, official sessions that count, by cell. Class from Table \ref{tab:ausencia} (product: from chain of thought on
the absent lengths). Validation: sealed-validation accuracy (\%) of each of the five sessions, as recorded by the operating
room; for tracking, the minimum over the model's absent subtasks, the unit of the preregistered rule. \emph{failed}: ended at the 12 h limit without an exported module; it counts as a session that
reaches no threshold; $^\dagger$: an infrastructure fault denied the session's validation request and we scored its exported
model on the same set (Appendix \ref{app:protocolo}). Passes: range over finished sessions. Neurons and heads: median added per module. Qwen 1.5B is
Qwen2.5-1.5B-Instruct and Gemma 1B is Gemma-3-1B-it, here and in Tables \ref{tab:test} and \ref{tab:modulos}.}
\label{tab:validacion}
\footnotesize
\setlength{\tabcolsep}{3.5pt}
\begin{tabularx}{\linewidth}{@{}lll>{\raggedright\arraybackslash}p{2.1cm}>{\raggedright\arraybackslash}Xrrr@{}}
\toprule
Task & Host & Manual & Class & Sealed validation, 5 sessions (\%) & Passes & Neurons & Heads\\
\midrule
\TabValidacion
\bottomrule
\end{tabularx}
\end{table}

\textbf{Results on the sealed test}\ExamenEstado. Every official module, both composites and the LoRA baseline (B1) of
each task were examined once by the frozen evaluator, after a one-time certificate with no violations (Table \ref{tab:test};
side effects of every module in Table \ref{tab:hdiez}; baselines in Table \ref{tab:honce}).

\emph{A 1.5B model above its 7B sibling.} The Qwen composite (\CCQwenN{} modules) scores \CCBBHQwen{} on the normalized BBH
score, against \CCCeroQwen{} for the base model as released (C0, bfloat16) and \CCCeroFQwen{} for its format control (C0-F: the
same base weights in float32, loaded through the same model file as the composites, with no module): a gain of \CCSubeQwen{} points (the
smaller of the two decides) that puts it above Qwen2.5-7B-Instruct in all three measurements (\BBHSieteOficial,
\BBHSieteLmeval{} and \BBHSieteComun). The Gemma composite (\CCGemmaN{} modules) goes from \CCCeroGemma{} (format control
\CCCeroFGemma) to \CCBBHGemma, a gain of \CCSubeGemma{} points over the format control. Against the format control, both have KL exactly 0 and no
top-token change on \FigUnoKLTokens{} WikiText tokens. \FraseCCPanel{} The gain therefore comes only from the six subtasks with a
module; the Qwen gain is close to the oracle defined above (\OraculoQwen{} points) and the Gemma gain reaches it
(\OraculoGemma{} points; the oracle was fixed before the exam from the lm-eval baseline, not from C0, so a gain can pass it):
the composites gain specific capabilities, not general ability. Against C0 in bfloat16, the composites differ only through
numerical precision: mean KL \CCKLCeroQwen{} (Qwen) and \CCKLCeroGemma{} (Gemma) and \CCTopCeroQwen{} and \CCTopCeroGemma{}
top-token changes per 10,000 WikiText tokens, the same as the format control against C0, which is why H10 measures every module
against the format control.

\emph{Module by module.} On fresh instances of the sealed test, tracking modules score \TestTrackQwen\% (3/5/7 objects, mean of
the five Qwen sessions; \HCincoSesQwen{} sessions at 90\% or more on every absent subtask, median \HCincoMedianaQwen\%) and
\TestTrackGemma\% in Gemma (mean of the examined sessions; \HCincoSesGemma{} at 90\% or more on every absent subtask, median
\HCincoMedianaGemma\%, the two failed sessions counting as 0). web\_of\_lies modules
score \TestWolQwen\% and \TestWolGemma\%, from \HOchoUnaQwen\% and \HOchoUnaGemma\% for the base model in one pass and
\HOchoCotQwen\% and \HOchoCotGemma\% with its best chain of thought. navigate and boolean\_expressions, designed without a task
manual, score \TestNavQwen\% and \TestBoolQwen\% in Qwen (\SesNavQwen{} and \SesBoolQwen{} sessions with a module; the
other \FallNavQwen{} and \FallBoolQwen{} reached the 12-hour limit and count as 0 in the rules) and \TestNavGemma\% and \TestBoolGemma\% in Gemma. Product modules score
\ProdTestQwen\% (Qwen) and \ProdTestGemma\% (Gemma) on the absent lengths and \ProdLejanasQwen\% and \ProdLejanasGemma\% from
\ProdLejanasDesde{} to \ProdLejanasHasta{} digits, more than twice the longest operand the agent could run. On the public BBH
items, never shown to the agent, the modules score as on our fresh instances (Table \ref{tab:test}).

\emph{Against a LoRA with the same budget.} B1 fits the template well: on fresh instances it matches the modules on most BBH
subtasks (for example \BUnoTestWolQwen\% and \BUnoTestWolGemma\% on web\_of\_lies). It falls behind on tracking with 5 and 7
objects in Gemma (\BUnoTestTrackGemma\% for 3/5/7 objects), on rewritten web\_of\_lies (\BUnoReescWolQwen\% and
\BUnoReescWolGemma\%, against \ReescWolQwen\% and \ReescWolGemma\% for the modules) and on multiplication at the absent lengths
(\BUnoProdAusQwen\% and \BUnoProdAusGemma\%). The comparison holds at the fixed budget of \Presupuesto{} passes; the total
cost of the agent, in passes and hours per session, is in Section \ref{sec:costo}. It also changes the model on unrelated text: its mean KL against the format
control is \BUnoKLQwen{} and \BUnoKLGemma{} nats per token, against \CCKLQwen{} for the composites.

\emph{What did not work.} Of the \HDiezModN{} single modules, \HDiezModFracaso{} fail H10 (\HDiezModFracasoLista): their
largest effects are \HDiezPeorTop{} top-token changes per 10,000 tokens (\HDiezPeorTopSesion; the limit is \HDiezTopUno) and a
drop of \HDiezPeorCaida{} points on \HDiezPeorCaidaPrueba{} (\HDiezPeorCaidaSesion), with every module's mean KL at most \HDiezPeorKL{}
(limit \HDiezKLMedia). None of the four entered a composite. The panel of \AnfitrionSesion, which changes one original matrix,
is not measured (Appendix \ref{app:protocolo}). The \DosFallidas{} failed sessions count in their cells as sessions that reach
no threshold; this is why the H8 sentence for tracking with 3 objects in Gemma below averages over five sessions, two of them at
0. On the rewritten test, \ReescModCien{} of the \ReescModN{} examined BBH sessions keep 100\% on every subtask; the lowest is
\ReescModMin\% (\ReescModMinSesion, \ReescModMinSub), and the Gemma composite scores \CCReescMinGemma\% on rewritten
\CCReescMinSubGemma, so H9 is partial in Gemma.

\emph{Preregistered verdicts:} \VeredictoQwen; \VeredictoGemma. \FrasePanelLento

\begin{framed}
\small\noindent\textbf{Preregistered verdict sentences.} One sentence per verdict that came out, labelled with its hypothesis and filled by
code from the verdict file. Sentences fixed in the sealed specification are reproduced word for word, in the wording of
amendment E2-X26 for H5 and H10 (iv); the Gemma H5 sentence, sealed as a tail of the Qwen one, is printed on its own. Sentences
marked $^\ast$ are descriptive: where the specification fixes none, code wrote them before the exam was gathered (Appendix
\ref{app:protocolo}).\par\noindent
\FrasesVeredicto{}
\end{framed}
\noindent In the H8 sentences, the final percentage is the mean over the five sessions of the cell, failed sessions counting as 0
(Gemma, tracking with 3 objects: three sessions at 99.8\% and two failed), and the one-pass value is that of the base model on
the sealed test. In the H11 sentences, failed sessions also count as 0; ``match'' allows the modules to be up to 2 points below B1 on the
own test, and ``beat'' also requires 10 points more than B1 on the rewritten test, which is why a cell of Table \ref{tab:test}
can be above B1 and still not count as beaten.

\begin{table}[t]
\centering
\caption{Stage 2 on the sealed test (preregistered)\ExamenEstado. Mean over the examined official sessions of each cell, in
\% (tracking: 3/5/7 objects; ($k$/5) when fewer than five were examined: failed sessions count as 0 in the decision rules).
Own test: fresh instances from our generator (product: its absent lengths, and lengths \ProdLejanasDesde{} to
\ProdLejanasHasta); public BBH: the \BBHPublicoN{} public items per subtask; rewritten: the sealed rewritten test; LoRA (B1):
the trained baseline with the same budget. Bottom: the composite model of each host, with its normalized BBH score, its gain
over the base model and over the format control (the smaller decides), its KL against the format control, and the rest of the
Leaderboard panel (H10 iii; no significant change: no test drops by more than 1 point and none is significant after Holm,
amendment E2-X26; for both composites, the same generation or choice as the format control on every non-BBH item and the same
outcome on every item of the other BBH subtasks); nested: the three-task
composites of amendment E2-X8, without H6 or H10.}
\label{tab:test}
\footnotesize
\setlength{\tabcolsep}{3pt}
\begin{tabular}{llllll}
\toprule
Task & Host & Own test & Public BBH & Rewritten & LoRA (B1)\\
\midrule
\TabTestDos
\midrule
Model & Host & BBH score & Gain & KL & Panel\\
\midrule
\TabTestCC
\bottomrule
\end{tabular}
\end{table}

\textbf{Do modules compose?} A module enters the composite model only if, composed with the modules already chosen, it
changes no prediction of the others on their practice sets; this rule was written after a development module whose gate opened
on every newline broke the tracking module. In development, three modules (tracking, web\_of\_lies and product) compose in
Qwen2.5-1.5B with the same predictions as alone (\CompDevMisma{} practice instances), and the composite answers
\CompDevCC\% of the public tracking-3 items against \CompDevBase\% for the base model ($n=\CompDevN$), with mean KL
$\CompDevKL$ and \CompDevCambios{} top-token changes per 10,000 tokens on \CompDevPasajes{} WikiText passages. For the official
composite of each host, the module of each task was chosen by rules fixed before composing: the highest sealed validation,
then the lowest mean KL against the format control on WikiText, measured outside the exam, then fewer passes; a candidate enters
only if no prediction changes on the practice set of any module of the composite, its own included; and the \AuditTareaN{}
modules whose marker direction was set on task inputs are not eligible (amendments E2-X1, E2-X7, E2-X10 and E2-X14; Appendix
\ref{app:auditoria}). The damage of each composite was predicted from its modules and sealed before composing (amendment
E2-X11, damage prediction). The
Qwen composite has \CCQwenN{} modules (\CCQwenLista) and the Gemma composite \CCGemmaN{} (\CCGemmaLista): in Gemma no
\CCGemmaFuera{} module composes with the tracking modules, because on the practice sets either the product module loses its own
answers inside the composite or tracking's predictions change. Both composites are examined once, like every module (Table
\ref{tab:test}). The nested three-task composites (\texttt{cc3-qwen15} and \texttt{cc3-gemma1}, amendment E2-X8) were also
examined once; H6 and H10 apply only to the composites of all tasks.

\textbf{Do modules survive new wording?} The sealed rewritten sets change how sentences are built, a family of constructions
announced to second-round surgeons whose instances they never saw. The rewritten test (Table \ref{tab:test}) therefore measures
generalization to fresh instances of an announced family of rewritings, not to free wording; the first-round development
module that motivated the second round is described in Appendix \ref{app:modulos}. On it, most modules keep their accuracy
(cell means in Table \ref{tab:test}; the drops are listed above), and Table \ref{tab:hnueve} breaks down each composite by
factor of the rewriting.

\section{Cost and the curve of the surgeon}
\label{sec:costo}

An implant costs \SumaPasadasMin{} to \SumaPasadasMax{} passes for addition and \DosPasadasMin{} to \DosPasadasMax{} for
stage-2 tasks (Tables \ref{tab:suma} and \ref{tab:validacion}); the LoRA baseline saw \LoraEjemplos{} examples, the equivalent
of the whole budget. From its first to its last call to the model, an addition session took a median of \SumaHorasMed{} hours
of wall-clock time (\SumaHorasMin{} to \SumaHorasMax) and a finished stage-2 session \DosHorasMed{} hours (\DosHorasMin{} to
\DosHorasMax). Table \ref{tab:curva} (Appendix \ref{app:curva}) compares development sessions with official sessions that use the frozen manual. The
official sessions reach their first exact implant on practice in fewer passes in \CurvaMejorEstricto{}, in as many in
\CurvaIgual{} and in more in \CurvaPeor{} of \CurvaN{} cells (more: \CurvaPeorCeldas). The gain is small, comes from what the
manual records, and has no concurrent control: we read it as the value of a written manual, not as a surgeon that learns.

\section{Related work}
\label{sec:related}

\textbf{Weight-space programming.} \citet{takehi2026sparks} showed that frontier models read and write network weights much like
code, and that a rank-one edit can remove a failure mode of a reasoning model, while noting that such edits uncover rather than
add skills. Follow-ups teach facts and an invented operator by editing Qwen weights without gradients \citep{zavo2026teaching},
compose independently found edits \citep{grover2026compose}, install new capabilities in small networks that are not pretrained
language models \citep{weightcode2026}, or let a model edit itself \citep{fan2026language}; others write a classifier into an
LLM layer by solving a regression on labelled examples \citep{tosmith2026banking}, which is fitting to task data.
\citet{marasini2026weightbench} find that agents do not tell a blocked capability from an absent one, which is why we measure
absence with chain of thought and with the algorithm in the prompt. Closest to our stage 1, an agent writes 3-digit addition into
the existing weights of GPT-2 small \citep{sudorian2026implants}, and agents write, without training, a 12-digit adder into ten
new layers placed before Qwen2.5-0.5B \citep{mfaseehuddin2026wsp}. We go further on three axes: the circuit lives in the existing layers (no added
depth; original tensors untouched), and off-trigger next-token distributions are identical in the session evaluated in full and
differ at \DanoDosCambios{} of \DanoPosiciones{} positions, all near-ties, in the second; it extrapolates from the
\SumaPracticaMax{} digits the agent could run to \SumaLongMax; and it is reproduced in \SumaPerfectas{} of \SumaOficiales{}
sessions on a sealed test.

\textbf{Modules and computers inside the forward pass.} \citet{granitzer2026tool} solve all 750 public tracking items in
Gemma-3-1B with a frozen tape module whose writer, reader and gate are trained; Turnstyle reaches 100\% on tracking and
web\_of\_lies with a parser and exact solver outside the weights \citep{donaldson2026swollm}. Compilers write programs into
transformer weights: Tracr \citep{lindner2023tracr}, a calculator compiled into a separate transformer
\citep{porter2026calculator}, a CPU compiled into new layers placed before Qwen3.5 and called through its tool-call format
\citep{qwenputer2026,concept2026rv32im}, and agent-written transformers built from scratch \citep{mehta2026interpretable}.
Neural Comprehension \citep{weng2024mastering} compiles rules written in RASP, by experts or by an LLM that writes Tracr code,
into separate modules that take over the output of a language model when their rule applies, with accuracy that does not
degrade with operand length; Tracr-injection \citep{vergara2025tracr} instead distills a compiled program into GPT-2 by
fine-tuning all its weights on task data, with a loss that aligns a projection of its residual stream to the compiled model.
Other work trains an interface to an exact adder inside an MLP \citep{wilson2026neurons} or adds exact arithmetic with LoRA and
an external runtime \citep{mohsenvand2026sketchpad}. Our modules are neither compiled nor trained or distilled: an agent writes
them as gated groups added inside the existing layers of a pretrained model, and they are judged on a sealed test.

\textbf{Length generalization and how trained models compute.} Trained transformers learn to extrapolate in length with
scratchpads in context \citep{anil2022exploring}, on tasks with short RASP-L programs \citep{zhou2024algorithms}, with Abacus
embeddings \citep{mcleish2024transformers} and with position coupling \citep{cho2024position}, and expressivity results
characterize what transformers compute in one forward pass \citep{merrill2023parallelism}. Interpretability shows how trained
models add with helical number representations \citep{kantamneni2025language} and why standard fine-tuning misses multi-digit
multiplication \citep{bai2025transformers}, and a language model can write programs that stand in for attention heads
\citep{hayes2026explaining}. Our work is complementary to all of them: here the circuit is written rather than learned, and the
agent's program goes into the weights, the direction opposite to that of \citet{hayes2026explaining}.

\textbf{Model editing and steering.} Locate-and-edit methods change stored facts \citep{meng2022locating}, and steering adds
activation vectors \citep{turner2023steering}; both reuse what a model already computes. We add computation it does not have.

\section{Limitations}
\label{sec:limits}

\begin{itemize}
\item \textbf{Templates.} Modules fire on the task template. Stage-2 rewritings test an announced family of constructions, and
  even there some sessions drop (to \ReescModMin\% in the worst session and subtask; the Gemma composite is at
  \CCReescMinGemma\% on rewritten \CCReescMinSubGemma); a probe of constructions never described to any surgeon, written by an
  agent that has not seen the manuals, is not part of this preprint. We do not claim generalization to free natural language.
  Modules depend on that template: what they add is the computation for the fixed task format, not an
  ability to read the task in any wording. Near forms of a task prompt can open a module's gates and change its output; the zero is measured on unrelated text
  (WikiText) and on the Leaderboard panel, not on every input.
\item \textbf{Exactness.} The addition implants have a finite carry window (Section \ref{sec:suma}): a long enough run of
  columns whose digits sum to 9 can make them fail. They are exact on the sealed test, not proven exact on every input.
\item \textbf{Scope.} Small hosts (0.5B to 1.5B), two families, one agent model, a handful of algorithmic tasks. The composite
  is above the 7B model on the BBH score because its six module subtasks reach the ceiling; on the other \PanelBBHSinModN{} subtasks and
  the rest of the panel it matches its format control (identical scores on those subtasks, identical answers elsewhere). The modules add these capabilities; we make no claim about reasoning in general.
\item \textbf{Side effects and failed sessions.} Of the \HDiezModN{} single modules, \HDiezModFracaso{} fail the
  side-effect rule (\HDiezModFracasoLista), and the panel of \AnfitrionSesion{} is not measured. Of the \DosPlanificadas{}
  stage-2 sessions, \DosFallidas{} reached the 12-hour cap without a module, after using only \FallPasadasMin{} to
  \FallPasadasMax{} of their \Presupuesto{} passes; they count as failures in every rule.
\item \textbf{The rule and its audit.} The method avoids backpropagation and fitting any weight to the answers of task
  examples, but some design constants
  come from small numerical searches over the model's geometry, its vocabulary or its activations (Section \ref{sec:cirujano},
  Appendix \ref{app:auditoria}).
\item \textbf{Validation is not test.} The agent could request validation accuracy per length or subtask, so validation
  scores are optimistic; \EstadoLimites. For stage 1, KL is measured in full in two sessions, MMLU and ARC in one; the stage-2
  exam measures KL on WikiText and the Leaderboard panel for every official module and both composites, except the panel of
  \AnfitrionSesion{} (Table \ref{tab:hdiez}).
\item \textbf{Rounds.} Tracking has a first official round whose development module did not survive rewriting; it was
  reclassified as development before the exam and its sealed test is never run, so the counted round benefits from a manual
  written after it.
\item \textbf{Hardware.} The exam runs on L4 GPUs and, for some parts and panel instances, on H200 GPUs (the own test and
  product of the LoRA baseline, on an H100). Parts were checked against the L4 on development implants: for Qwen before use,
  for Gemma after its H200 parts had run and never on the H100; panel instances on an H200 were never compared with the L4,
  only with base-model trajectories computed on the same H200 (Table \ref{tab:gpus}). The rented L4 GPUs are
  identical bit for bit on the GPU path, while the float64 CPU path of exact
  components, used only beyond 2,048 tokens of context, can differ in the last float32 bit on hosts without AVX-512; we report
  which session ran where. Implants are validated on L4 GPUs with one fixed software stack; on a
  laptop GPU (MLX on Apple Metal), \FigUnoMacFallan{} of the \FigUnoMacN{} Qwen product implants returned the base model's answer
  on the example of Figure \ref{fig:one}a, so we make no claim for other backends.
\item \textbf{Cost.} Sessions take hours of agent time; the parameter counts in our tables are dense allocations, and the
  number of nonzero weights is smaller.
\end{itemize}

\section{Future work}
\label{sec:futuro}

The results point to several directions, none measured in this paper. \emph{Capabilities that are not described to the agent}:
here the brief always says what the module must compute; the next step gives the agent only the task format and practice data.
\emph{Larger models}: the hosts here have at most 1.5B parameters; whether the same tools and budget suffice for wider and
deeper hosts is open. \emph{Robustness to rewriting}: modules that keep their accuracy on wording no surgeon was told about,
judged on a sealed set of free rewritings. \emph{Composing many modules}: our composites hold four or five
modules; a library of many modules needs rules that scale, and the product module in Gemma shows that composition can fail.
\emph{Removing the time cap}: the failed sessions stopped at 12 hours with most of their budget unused. \emph{Checking side
effects automatically}: the operating room could run the Leaderboard panel before export, so that a module like the four above
is caught by the agent itself. \emph{Interpreting the written weights}: every module comes with the code that wrote it, so the
circuits can be read back and compared with what the manual and the agent's report say they compute.

\section{Conclusion}

An agent with tools can write capabilities into a language model as new weights inside its existing layers, without
backpropagation and without fitting any weight to the answers of task examples. What it writes extrapolates beyond what it could
run (addition to \SumaLongMax{} digits, multiplication to \ProdLongMax), holds on sealed tests in two model families, and
composes into one model that matches the unmodified model at the same precision on everything else we measured.
\EstadoConclusion{} A 1.5B model with agent-written modules scores \CCBBHQwen{} on BIG-Bench Hard, where its 7B sibling scores
at most \BBHSieteComun{} in three measurements, and matches its format control on every other test we ran: some capabilities need not be
trained; they can be written.

\section*{Availability}
\Disponibilidad

\section*{AI use statement}
Generative AI is part of the method: the surgeon is Claude Code with an Anthropic model (Section \ref{sec:cirujano}). The author
also used Claude Code as a research assistant for literature search and novelty checks, for writing and debugging the code of the
operating room, evaluator and baselines, for running and monitoring experiments, for analysis scripts, and for drafting the text
of this manuscript, its tables and its figures. The research question and the decision to pursue it are the author's. To verify
AI-assisted outputs, every number in the paper is generated by a script from result files; sets, rules and the evaluator were
sealed or frozen before measuring; how each implanted weight was computed was audited by a separate agent against a criterion
fixed in advance; and every reference was checked against its source. No AI system is an author. The author takes
full responsibility for all content.

\bibliographystyle{plainnat}
\bibliography{references}

\begin{thebibliography}{43}
\providecommand{\natexlab}[1]{#1}
\providecommand{\url}[1]{\texttt{#1}}
\expandafter\ifx\csname urlstyle\endcsname\relax
  \providecommand{\doi}[1]{doi: #1}\else
  \providecommand{\doi}{doi: \begingroup \urlstyle{rm}\Url}\fi

\bibitem[Agrawal(2026)]{qwenputer2026}
Divyansh Agrawal.
\newblock {Qwenputer}: a computer compiled into the weights of a pretrained language model.
\newblock GitHub repository, version 0.1.0, \url{https://github.com/divagr18/qwenputer}, 2026.

\bibitem[Anil et~al.(2022)Anil, Wu, Andreassen, Lewkowycz, Misra, Ramasesh, Slone, Gur-Ari, Dyer, and Neyshabur]{anil2022exploring}
Cem Anil, Yuhuai Wu, Anders Andreassen, Aitor Lewkowycz, Vedant Misra, Vinay Ramasesh, Ambrose Slone, Guy Gur-Ari, Ethan Dyer, and Behnam Neyshabur.
\newblock {Exploring Length Generalization in Large Language Models}.
\newblock In \emph{Advances in Neural Information Processing Systems 35}, NeurIPS 2022, page 38546–38556. Neural Information Processing Systems Foundation, Inc. (NeurIPS), 2022.
\newblock \doi{10.52202/068431-2793}.
\newblock URL \url{http://dx.doi.org/10.52202/068431-2793}.

\bibitem[{Anthropic}(2026)]{anthropic2026claudecode}
{Anthropic}.
\newblock {Claude Code}.
\newblock Software, 2026.

\bibitem[Bai et~al.(2025)Bai, Pres, Deng, Tan, Shieber, Viégas, Wattenberg, and Lee]{bai2025transformers}
Xiaoyan Bai, Itamar Pres, Yuntian Deng, Chenhao Tan, Stuart Shieber, Fernanda Viégas, Martin Wattenberg, and Andrew Lee.
\newblock {Why Can't Transformers Learn Multiplication? Reverse-Engineering Reveals Long-Range Dependency Pitfalls}, 2025.
\newblock URL \url{https://arxiv.org/abs/2510.00184}.

\bibitem[Biderman et~al.(2024)Biderman, Schoelkopf, Sutawika, et~al.]{biderman2024lessons}
Stella Biderman, Hailey Schoelkopf, Lintang Sutawika, et~al.
\newblock {Lessons from the Trenches on Reproducible Evaluation of Language Models}, 2024.
\newblock URL \url{https://arxiv.org/abs/2405.14782}.

\bibitem[Cho et~al.(2024)Cho, Cha, Awasthi, Bhojanapalli, Gupta, and Yun]{cho2024position}
Hanseul Cho, Jaeyoung Cha, Pranjal Awasthi, Srinadh Bhojanapalli, Anupam Gupta, and Chulhee Yun.
\newblock {Position Coupling: Improving Length Generalization of Arithmetic Transformers Using Task Structure}.
\newblock In \emph{Advances in Neural Information Processing Systems 37}, NeurIPS 2024, page 22233–22315. Neural Information Processing Systems Foundation, Inc. (NeurIPS), 2024.
\newblock \doi{10.52202/079017-0701}.
\newblock URL \url{http://dx.doi.org/10.52202/079017-0701}.

\bibitem[Clark et~al.(2018)Clark, Cowhey, Etzioni, Khot, Sabharwal, Schoenick, and Tafjord]{clark2018think}
Peter Clark, Isaac Cowhey, Oren Etzioni, Tushar Khot, Ashish Sabharwal, Carissa Schoenick, and Oyvind Tafjord.
\newblock {Think you have Solved Question Answering? Try ARC, the AI2 Reasoning Challenge}, 2018.
\newblock URL \url{https://arxiv.org/abs/1803.05457}.

\bibitem[{Concept Country}(2026)]{concept2026rv32im}
{Concept Country}.
\newblock {Exact Neural Network Inference Inside the Weights of Qwen Concept 27B}.
\newblock Working paper v1, \url{https://concept.country/research/concept-rv32im.pdf}, 2026.
\newblock November 8.

\bibitem[Donaldson(2026)]{donaldson2026swollm}
Justin Donaldson.
\newblock {A 1.7B Model That Stops Guessing}.
\newblock Blog post, \url{https://jjd.io/posts/swollm-bbh-leaderboard.html}, 2026.
\newblock June 27.

\bibitem[Fan et~al.(2026)Fan, He, Yang, et~al.]{fan2026language}
Jiajun Fan, Neil He, Changyi Yang, et~al.
\newblock {Language Models Can Rewrite Their Own Weights}, 2026.
\newblock URL \url{https://arxiv.org/abs/2611.21885}.

\bibitem[Fourrier et~al.(2024)Fourrier, Habib, Lozovskaya, Szafer, and Wolf]{fourrier2024leaderboard}
Cl{\'e}mentine Fourrier, Nathan Habib, Alina Lozovskaya, Konrad Szafer, and Thomas Wolf.
\newblock {Open LLM Leaderboard v2}.
\newblock Hugging Face, \url{https://huggingface.co/spaces/open-llm-leaderboard/open_llm_leaderboard}, 2024.

\bibitem[Gao et~al.(2023)Gao, Madaan, Zhou, Alon, Liu, Yang, Callan, and Neubig]{gao2023pal}
Luyu Gao, Aman Madaan, Shuyan Zhou, Uri Alon, Pengfei Liu, Yiming Yang, Jamie Callan, and Graham Neubig.
\newblock {PAL: Program-aided Language Models}.
\newblock In \emph{International Conference on Machine Learning (ICML)}, 2023.
\newblock URL \url{https://openreview.net/forum?id=M1fd9Z00sj}.

\bibitem[{Gemma Team} et~al.(2025){Gemma Team}, Kamath, Ferret, et~al.]{team2025gemma}
{Gemma Team}, Aishwarya Kamath, Johan Ferret, et~al.
\newblock {Gemma 3 Technical Report}, 2025.
\newblock URL \url{https://arxiv.org/abs/2503.19786}.

\bibitem[{Gen-Verse}(2026)]{weightcode2026}
{Gen-Verse}.
\newblock {WeightCode: Installing New Capabilities into Neural Networks via Weight-Space Programming}.
\newblock GitHub repository, \url{https://github.com/Gen-Verse/WeightCode}, 2026.

\bibitem[Granitzer et~al.(2026)Granitzer, Zerhoudi, Al-Maamari, Dang, and Dastidar]{granitzer2026tool}
Michael Granitzer, Saber Zerhoudi, Mohammad Al-Maamari, Hai~Dang Dang, and Kanishka~Ghosh Dastidar.
\newblock {A Tool in the Forward Pass: Frozen Tape Modules as Layers of a Small Language Model}, 2026.
\newblock URL \url{https://arxiv.org/abs/2611.00783}.

\bibitem[Grover(2026)]{grover2026compose}
Ajay Grover.
\newblock {Independently Discovered Weight Edits Compose}.
\newblock Blog post, \url{https://ajaygrover.github.io/weight-space-programming-stacking/}, 2026.
\newblock November 7.

\bibitem[Hansen(2016)]{hansen2016cma}
Nikolaus Hansen.
\newblock {The CMA Evolution Strategy: A Tutorial}, 2016.
\newblock URL \url{https://arxiv.org/abs/1604.00772}.

\bibitem[Hayes et~al.(2026)Hayes, Li, and Andreas]{hayes2026explaining}
Amiri Hayes, Belinda~Z Li, and Jacob Andreas.
\newblock {Explaining Attention with Program Synthesis}, 2026.
\newblock URL \url{https://arxiv.org/abs/2606.19317}.

\bibitem[Hendrycks et~al.(2020)Hendrycks, Burns, Basart, Zou, Mazeika, Song, and Steinhardt]{hendrycks2020measuring}
Dan Hendrycks, Collin Burns, Steven Basart, Andy Zou, Mantas Mazeika, Dawn Song, and Jacob Steinhardt.
\newblock {Measuring Massive Multitask Language Understanding}, 2020.
\newblock URL \url{https://arxiv.org/abs/2009.03300}.

\bibitem[Hu et~al.(2021)Hu, Shen, Wallis, Allen-Zhu, Li, Wang, Wang, and Chen]{hu2021lora}
Edward~J. Hu, Yelong Shen, Phillip Wallis, Zeyuan Allen-Zhu, Yuanzhi Li, Shean Wang, Lu~Wang, and Weizhu Chen.
\newblock {LoRA: Low-Rank Adaptation of Large Language Models}, 2021.
\newblock URL \url{https://arxiv.org/abs/2106.09685}.

\bibitem[Kantamneni and Tegmark(2025)]{kantamneni2025language}
Subhash Kantamneni and Max Tegmark.
\newblock {Language Models Use Trigonometry to Do Addition}, 2025.
\newblock URL \url{https://arxiv.org/abs/2502.00873}.

\bibitem[Lindner et~al.(2023)Lindner, Kramar, Farquhar, Rahtz, McGrath, and Mikulik]{lindner2023tracr}
David Lindner, Janos Kramar, Sebastian Farquhar, Matthew Rahtz, Tom McGrath, and Vladimir Mikulik.
\newblock {Tracr: Compiled Transformers as a Laboratory for Interpretability}.
\newblock In \emph{Advances in Neural Information Processing Systems 36}, NeurIPS 2023, page 37876–37899. Neural Information Processing Systems Foundation, Inc. (NeurIPS), 2023.
\newblock \doi{10.52202/075280-1649}.
\newblock URL \url{http://dx.doi.org/10.52202/075280-1649}.

\bibitem[Marasini et~al.(2026)Marasini, Khadka, and Rahimi]{marasini2026weightbench}
Aayush Marasini, Barsat Khadka, and Nick Rahimi.
\newblock {WeightBench: Evaluating Language Models as Weight-Space Programmers}, 2026.
\newblock URL \url{https://arxiv.org/abs/2611.18553}.

\bibitem[McLeish et~al.(2024)McLeish, Bansal, Stein, Jain, Kirchenbauer, Bartoldson, Kailkhura, Bhatele, Geiping, Schwarzschild, and Goldstein]{mcleish2024transformers}
Sean McLeish, Arpit Bansal, Alex Stein, Neel Jain, John Kirchenbauer, Brian Bartoldson, Bhavya Kailkhura, Abhinav Bhatele, Jonas Geiping, Avi Schwarzschild, and Tom Goldstein.
\newblock {Transformers Can Do Arithmetic with the Right Embeddings}.
\newblock In \emph{Advances in Neural Information Processing Systems 37}, NeurIPS 2024, page 108012–108041. Neural Information Processing Systems Foundation, Inc. (NeurIPS), 2024.
\newblock \doi{10.52202/079017-3430}.
\newblock URL \url{http://dx.doi.org/10.52202/079017-3430}.

\bibitem[Mehta et~al.(2026)Mehta, Singh, Xu, et~al.]{mehta2026interpretable}
Yash Mehta, Chandan Singh, Weijia Xu, et~al.
\newblock {Interpretable weight-space programming builds transparent transformers and brain encoding models}, 2026.
\newblock URL \url{https://arxiv.org/abs/2611.18114}.

\bibitem[Meng et~al.(2022)Meng, Bau, Andonian, and Belinkov]{meng2022locating}
Kevin Meng, David Bau, Alex Andonian, and Yonatan Belinkov.
\newblock {Locating and Editing Factual Associations in GPT}.
\newblock In \emph{Advances in Neural Information Processing Systems 35}, NeurIPS 2022, page 17359–17372. Neural Information Processing Systems Foundation, Inc. (NeurIPS), 2022.
\newblock \doi{10.52202/068431-1262}.
\newblock URL \url{http://dx.doi.org/10.52202/068431-1262}.

\bibitem[Merity et~al.(2016)Merity, Xiong, Bradbury, and Socher]{merity2016pointer}
Stephen Merity, Caiming Xiong, James Bradbury, and Richard Socher.
\newblock {Pointer Sentinel Mixture Models}, 2016.
\newblock URL \url{https://arxiv.org/abs/1609.07843}.

\bibitem[Merrill and Sabharwal(2023)]{merrill2023parallelism}
William Merrill and Ashish Sabharwal.
\newblock {The Parallelism Tradeoff: Limitations of Log-Precision Transformers}.
\newblock \emph{Transactions of the Association for Computational Linguistics}, 11:\penalty0 531–545, 2023.
\newblock ISSN 2307-387X.
\newblock \doi{10.1162/tacl_a_00562}.
\newblock URL \url{http://dx.doi.org/10.1162/tacl_a_00562}.

\bibitem[{mfaseehuddin}(2026)]{mfaseehuddin2026wsp}
{mfaseehuddin}.
\newblock {wsp, Track A: bionic circuits}.
\newblock GitHub repository, \url{https://github.com/mfaseehuddin/wsp}, 2026.
\newblock November 6 to 8.

\bibitem[Mohsenvand(2026)]{mohsenvand2026sketchpad}
Mostafa~{`}Neo{'} Mohsenvand.
\newblock {Sketchpad: certified exact arithmetic with a small vision-language model}, 2026.
\newblock URL \url{https://arxiv.org/abs/2611.04434}.

\bibitem[Porter(2026)]{porter2026calculator}
Rob Porter.
\newblock {A calculator, compiled into a transformer}.
\newblock Blog post, \url{https://ood.dev/posts/calculator/}, 2026.
\newblock August 9.

\bibitem[{Qwen Team} et~al.(2024){Qwen Team}, Yang, Yang, et~al.]{qwen2024qwen2}
{Qwen Team}, An~Yang, Baosong Yang, et~al.
\newblock {Qwen2.5 Technical Report}, 2024.
\newblock URL \url{https://arxiv.org/abs/2412.15115}.

\bibitem[Saxton et~al.(2019)Saxton, Grefenstette, Hill, and Kohli]{saxton2019analysing}
David Saxton, Edward Grefenstette, Felix Hill, and Pushmeet Kohli.
\newblock {Analysing Mathematical Reasoning Abilities of Neural Models}, 2019.
\newblock URL \url{https://arxiv.org/abs/1904.01557}.

\bibitem[{sfc-gh-tosmith}(2026)]{tosmith2026banking}
{sfc-gh-tosmith}.
\newblock {Banking77 by weight-space programming}.
\newblock GitHub repository, \url{https://github.com/sfc-gh-tosmith/wsp-banking77}, 2026.

\bibitem[{sudorian}(2026)]{sudorian2026implants}
{sudorian}.
\newblock {Weight-space implants}.
\newblock GitHub repository, \url{https://github.com/sudorian/weight-space-implants}, 2026.
\newblock Created November 10.

\bibitem[Suzgun et~al.(2023)Suzgun, Scales, Schärli, et~al.]{suzgun2023challenging}
Mirac Suzgun, Nathan Scales, Nathanael Schärli, et~al.
\newblock {Challenging BIG-Bench Tasks and Whether Chain-of-Thought Can Solve Them}.
\newblock In \emph{ACL (Findings)}, 2023.
\newblock URL \url{https://openreview.net/forum?id=3q0IqOJgEB}.

\bibitem[Takehi et~al.(2026)Takehi, Teoh, Zhang, and Khattab]{takehi2026sparks}
Rikiya Takehi, Jayden Teoh, Alex Zhang, and Omar Khattab.
\newblock {Sparks of Weight-Space Programming: LLMs can natively read \& write neural network weights}.
\newblock Blog post, MIT CSAIL, \url{https://rikiyat.github.io/blog/weight-space-programming/}, 2026.
\newblock November 2.

\bibitem[Turner et~al.(2023)Turner, Thiergart, Leech, Udell, Vazquez, Mini, and MacDiarmid]{turner2023steering}
Alexander~Matt Turner, Lisa Thiergart, Gavin Leech, David Udell, Juan~J. Vazquez, Ulisse Mini, and Monte MacDiarmid.
\newblock {Steering Language Models With Activation Engineering}, 2023.
\newblock URL \url{https://arxiv.org/abs/2308.10248}.

\bibitem[Vergara~Browne and Soto(2025)]{vergara2025tracr}
Tom{\'a}s Vergara~Browne and Alvaro Soto.
\newblock {Tracr-Injection}: Distilling algorithms into pre-trained language models.
\newblock In \emph{Findings of the {A}ssociation for {C}omputational {L}inguistics: {ACL} 2025}, pages 2831--2843, Vienna, Austria, 2025. Association for Computational Linguistics.
\newblock \doi{10.18653/v1/2025.findings-acl.146}.
\newblock URL \url{https://aclanthology.org/2025.findings-acl.146/}.

\bibitem[Weng et~al.(2024)Weng, Zhu, Xia, Li, He, Liu, and Zhao]{weng2024mastering}
Yixuan Weng, Minjun Zhu, Fei Xia, Bin Li, Shizhu He, Kang Liu, and Jun Zhao.
\newblock {Mastering Symbolic Operations: Augmenting Language Models with Compiled Neural Networks}.
\newblock In \emph{International Conference on Learning Representations (ICLR)}, 2024.
\newblock URL \url{https://openreview.net/forum?id=9nsNyN0vox}.

\bibitem[Wilson(2026)]{wilson2026neurons}
Josiah Wilson.
\newblock {Deterministic ``Neurons'' Inside an Existing Transformer MLP}.
\newblock \url{https://josiahwilson.me/documents/neural-firmware-phase-7.pdf}, 2026.
\newblock July 27.

\bibitem[{Zavo}(2026)]{zavo2026teaching}
{Zavo}.
\newblock {Teaching a language model new knowledge by editing its weights}.
\newblock \url{https://zavo.ai/research/weight-editing}, 2026.
\newblock November 3.

\bibitem[Zhou et~al.(2024)Zhou, Bradley, Littwin, Razin, Saremi, Susskind, Bengio, and Nakkiran]{zhou2024algorithms}
Hattie Zhou, Arwen Bradley, Etai Littwin, Noam Razin, Omid Saremi, Josh Susskind, Samy Bengio, and Preetum Nakkiran.
\newblock {What Algorithms can Transformers Learn? A Study in Length Generalization}.
\newblock In \emph{International Conference on Learning Representations (ICLR)}, 2024.
\newblock URL \url{https://openreview.net/forum?id=AssIuHnmHX}.

\end{thebibliography}

\clearpage
\appendix

\section{Per-module details of stage 2}
\label{app:modulos}

\begingroup
\small
\setlength{\LTcapwidth}{\textwidth}
\begin{longtable}{lllrrrrrr}
\caption{Top: every finished stage-2 official session that counts, with its sealed-validation accuracy (\%; tracking:
minimum over the model's absent subtasks; $^\dagger$: scored by us, as in Table \ref{tab:validacion}), passes, and what the
final implant adds (from the patch headers; entries are dense allocations in millions). Bottom: tracking, round 1 (manual v3.1, \RondaUnoN{} sessions), not counted: reclassified as
development before the exam and never evaluated on any sealed test (amendment of 18 November); \RondaUnoSinVal{} requested no
validation and wrote no report.}
\label{tab:modulos}\\
\toprule
Session & Task & Host & Val. & Passes & Neurons & Heads & Layers & Entries (M)\\
\midrule
\endfirsthead
\multicolumn{9}{l}{\textit{Table \ref{tab:modulos}, continued}}\\
\toprule
Session & Task & Host & Val. & Passes & Neurons & Heads & Layers & Entries (M)\\
\midrule
\endhead
\bottomrule
\endfoot
\bottomrule
\endlastfoot
\TabModulos
\midrule
\multicolumn{9}{l}{\textit{Tracking, round 1 (manual v3.1): not counted}}\\
\TabRondaUno
\end{longtable}
\endgroup

\textbf{Why tracking has a second round.} A development tracking module of the first round, exact on its own variants, scored
\ReescDevTres\%, \ReescDevCinco\% and \ReescDevSiete\% on the sealed rewritten validation for 3, 5 and 7 objects, against
\ReescBaseTres\%, \ReescBaseCinco\% and \ReescBaseSiete\% for the base model, measured by us on the exported model. Its own
variants changed names, objects and single words within the same sentence construction; the rewritten set changes the
construction. The second-round manual says so (Appendix \ref{app:protocolo}), and only the second round counts.

\textbf{Side effects of every module (H10) and the comparison with baselines (H11).} The specification requires both tables
whatever the outcome; they are generated from the verdict files of the exam.

\begingroup
\scriptsize
\setlength{\tabcolsep}{3pt}
\setlength{\LTcapwidth}{\textwidth}
\begin{longtable}{@{}lllrrr>{\raggedright\arraybackslash}p{3.3cm}l@{}}
\caption{H10 on the sealed exam, per composite and per official module: mean KL against the format control on WikiText, its
99.9th percentile and top-token changes per 10,000 tokens (decision limits: \HDiezKLMedia, \HDiezPNueve{} and \HDiezTopUno), and the rest
of the Leaderboard panel (23 tests; no significant change: none drops by more than 1 point and none is significant after Holm;
not measured: with its cause and amendment). The negative mean KL of \texttt{of-prod-g02} is floating-point rounding of a mean below $10^{-9}$; its 99.9th percentile and top-token changes are not zero. Failed sessions count in
the rules and have no H10. (iv), the prediction of the
composite's damage, is in the verdict line of Section \ref{sec:bbh}.}
\label{tab:hdiez}\\
\toprule
Session & Task & Host & KL & p99.9 & Top-1 & Panel & H10\\
\midrule
\endfirsthead
\multicolumn{8}{l}{\textit{Table \ref{tab:hdiez}, continued}}\\
\toprule
Session & Task & Host & KL & p99.9 & Top-1 & Panel & H10\\
\midrule
\endhead
\bottomrule
\endlastfoot
\TabHDiez
\end{longtable}
\endgroup

\begin{table}[H]
\centering
\caption{H11 on the sealed exam: modules (M, mean of the examined official sessions of the cell), the composite (CC) and the
LoRA baseline with the same budget (B1), on fresh instances (own test) and on the rewritten test, in \%, and their damage (mean
KL against the format control on WikiText). B5 is the exact solver run as an external program, B7 a parser bound to the BBH
template and B8 the base model reasoning step by step.}
\label{tab:honce}
\footnotesize
\setlength{\tabcolsep}{4pt}
\begin{tabular}{@{}llrrrrrr@{}}
\toprule
Host & Subtask & M own & M rewr. & CC own & CC rewr. & B1 own & B1 rewr.\\
\midrule
\TabHOnce
\bottomrule
\end{tabular}
\end{table}

\begingroup
\footnotesize
\setlength{\LTcapwidth}{\textwidth}
\begin{longtable}{@{}ll>{\raggedright\arraybackslash}p{10.2cm}@{}}
\caption{H9, accuracy (\%) of each composite on the sealed rewritten test by factor of the rewriting, per subtask.}
\label{tab:hnueve}\\
\toprule
Host & Factor & Accuracy per subtask\\
\midrule
\endfirsthead
\toprule
Host & Factor & Accuracy per subtask\\
\midrule
\endhead
\bottomrule
\endlastfoot
\TabHNueve
\end{longtable}
\endgroup

\clearpage
\section{The curve of the surgeon}
\label{app:curva}

Table \ref{tab:curva} gives, per cell, the passes until the first exact implant on practice that Section \ref{sec:costo}
discusses.

\begin{table}[H]
\centering
\caption{Passes until the first implant with 100\% on practice: median over the sessions that got there (in parentheses: how
many got there of how many ran). Development sessions precede the frozen manual; official sessions use the manual version that
counts for that cell. Computed from the process files of each session only.}
\label{tab:curva}
\small
\begin{tabular}{lrlr}
\toprule
Task (host) & Development & Manual & Official\\
\midrule
\TabCurva
\bottomrule
\end{tabular}
\end{table}

\section{Audit of the rule on how weights are computed}
\label{app:auditoria}

The audit criterion was committed before any session folder was opened. For each session the auditor followed the active chain
of patches back to the scripts and files that produced each tensor and classified the session: A, closed-form linear algebra
(class means, linear discriminants, projections, exact minimum-norm solves of design constraints) over activations of instances
the agent writes itself; B, a numerical optimizer whose numbers are in the evaluated weights; C, such an optimizer tried and
discarded. Result over \AuditN{} sessions (\AuditDos{} of stage 2 and \AuditUno{} of stage 1): \AuditA{} A, \AuditB{} B and
\AuditC{} C. In \AuditTareaN{} sessions (\AuditTareaSesiones), the direction of a marker that makes a host attention head attend
to the task signature was set by gradient ascent on an attention margin over host activations on task inputs (practice items and
the agent's own instances), without using their answers: the objective is where the head looks, not the answer. These
\AuditTareaN{} modules do not enter the composite models (amendment E2-X14, written before any navigate or
boolean\_expressions module was chosen for a composite). The other B cases optimize over rotary geometry, the vocabulary or unrelated
text (Section \ref{sec:cirujano} lists the searches); one also chose the signs of two register directions by random draws that
minimize interference on activations that include practice items. The audit table released with the code gives the class and the searches of every session, including the \AuditGrad{} with
(sub)gradient ascent. No session counts differently because of the audit.

\section{Protocol details and amendments}
\label{app:protocolo}

\textbf{Budget.} \Presupuesto{} passes per session; a pass is a scoring or generation call of up to 512 prompts or an activation
read of up to 1,024 texts; requesting validation costs one pass. \textbf{Decision rules.} For example, the stage-2 rule for
absent tracking is success if in at least 3 of 5 sessions every absent subtask of the model reaches 90\% on the sealed test;
the composed model succeeds if its BBH gain reaches 70\% of the oracle gain of its modules, measured against the base model and
against the format control, with the smaller of the two deciding. \textbf{Amendments that matter here.} The B1 baseline also
tries the four LoRA shapes with a learning rate ten times smaller, because the inherited recipe diverged; the module of each
task in the composite model is the composable official module with the best validation, with ties broken by KL against the
format control; one round per cell counts, fixed before the exam; an official session with no validation score because of an
infrastructure fault is scored by us on validation only. \VistasN{} times an accidental view showed exam accuracies: on 18
November at \VistoHora{} CET, \VistoN{} instances of the test of tracking with 3 objects of \VistoSesion{} (amendment E2-X3); at
\VistoDosHora{} CET, \VistoDosN{} instances of the rewritten test of tracking with \VistoDosObjetos{} objects of
\VistoDosSesion{} (E2-X12); on 19 November at \VistoTresHora{} CET, two log lines with the accuracy of the Qwen composite on one
subtask of the rewritten test, read by an agent and never by the author (E2-X19). The sessions that count, the evaluator, the
sealed sets and the decision rules of the specification were fixed before all three. The rule that picks each composite's
modules (E2-X1) was written after the first view: among \CCTrackEmpatadas{} tracking modules tied at 100\% validation and zero
KL it takes the one with the fewest passes, which is \VistoSesion, the session of that view. The amendments written after
the third view (from E2-X20 on) only say where and how exam parts and panel instances were computed, how an unfinished panel is
reported, in E2-X26 how two sentences are worded and which go into the abstract, and in E2-X28 how the data of each machine
are kept; none changes a set, a threshold or a decision rule in a direction that could favour a result. E2-X29 measures, once
and before the exam was gathered, the third 7B reference that H6 already required: without it, H6 could never call a composite
above its 7B sibling, and, being the highest of the three, it asks more than the other two alone. The second-round manual told surgeons the
aggregate scores of one development module on the sealed rewritten validation (\ReescDevTres, \ReescDevCinco{} and
\ReescDevSiete\%) and that the rewritten set changes sentence construction; its instances were never shown. Each amendment is
dated and hashed before the exam was gathered (Section \ref{sec:protocol}). Two amendments share the number E2-X11 (of-wol-q05,
and the damage prediction of the composites, both of 18 November); E2-X9 was numbered while its GPU check ran and sealed at
\XNueveSellada{} CET, after E2-X10 to E2-X13 (E2-X13 at \XTreceSellada).

\textbf{Wording of two preregistered sentences (amendment E2-X26).} The sealed sentence for H5 said the surgeon writes the
capability ``without gradients or training data'', and the one for H10 (iv) said the composite's side effects were predicted
``from each module's gate sweep''. The audit of Section \ref{sec:cirujano} found small numerical searches in final weights, and
the prediction actually sealed (E2-X11, damage prediction) adds each module's KL on WikiText and assumes no effect on the rest
of the panel. In an amendment sealed before the exam was gathered and written without knowledge of any verdict
(\XVeintiSeisHora{} CET on 19 November), the paper's wording became ``without backpropagation and without fitting any weight to the answers of task examples'' and ``from
each module's KL on unrelated text and a null effect on the rest of the panel''; the verdict files keep the sealed wording. The
same amendment fixed which exam sentences go into the abstract (H5 in Qwen, then H6 and H12 in each model, each in the version
of its verdict), so that none is chosen after seeing the result.

\textbf{Presentation after the exam (amendment E2-X30).} After the exam was gathered, and only for presentation, amendment
E2-X30 changed only presentation: no set, threshold, rule or verdict. The abstract had grown
beyond the 1,920 characters that arXiv accepts and, against the letter of E2-X26 (2), it carried sentences about the exam
written after seeing it. Under E2-X30 the abstract has a free summary, written after the exam from the result files and true
for every result we report, and the mechanical list of every verdict of both models, chosen by no one (including the partial
H9 in Gemma); the sentence of every verdict goes once in the box ``Preregistered verdict
sentences'' of Section \ref{sec:bbh}: the sealed ones word for word and, where the specification fixes none (H11 against
LoRA, and H12 in Gemma), the descriptive one written by code before the exam was gathered. The free text of the introduction and the conclusion was also written after the exam,
with every number taken from the result files. In the sealed tail of H6 (``above its 7B sibling (Y), measured with the same
protocol''), Y now names whose score each number is (the composite's, then the three of the 7B), because the composite's score
alone read as the 7B's; that tail joins its sentence with a comma, a form also set after the exam. The panel sentence of the
composites says ``same generation or choice'' only where answers were compared item by item and ``same outcome'' on the BBH subtasks, where
the evaluator records only whether each answer is correct.

\textbf{The 7B reference of H6 (amendment E2-X29).} H6 calls a composite above its 7B sibling only if its normalized BBH score
exceeds the official one of Qwen2.5-7B-Instruct, ours with lm-eval-harness and, as the sealed specification says, the one with our
common evaluator once it exists. That third score did not exist. In an amendment written at \XVeintiNueveHora{} CET on 19 November,
before measuring it and without seeing any exam result, we fixed how it is measured: the public checkpoint, whose weight files
match the SHA-256 published by Hugging Face, through the same frozen evaluator and the same path as the public BBH part of the
base model, in bfloat16 on an L4 GPU (the GPU type of that part), once, its value counting whatever it was. It ran from
\BBHSieteDesde{} to \BBHSieteHasta{} CET and gave \BBHSieteComun{} on the \BBHSieteComunN{} items of the 24 subtasks, against
\BBHSieteOficial{} official and \BBHSieteLmeval{} with lm-eval-harness.

\textbf{Exceptions and incidents.} The panel of \AnfitrionSesion, the only implant that changes an original matrix, is not
measured: with its implant outputs set to zero it does not reproduce the format control, so its panel has no valid reference
(amendments E2-X11 and E2-X18). Its KL and top-token changes against the format control are reported as measured, and its H10
never counts as a success. Paired comparisons (a module against the base model, the format control or the LoRA baseline) use
the same evaluator and numerical type, but not always the same machine or GPU type, as our specification first required
(assumption A15): RunPod and Google L4 GPUs are identical bit for bit on the GPU path, and on H100 and H200 GPUs we checked
predictions, not log-likelihoods (Table \ref{tab:gpus}); we publish the machine and GPU of each side. \FraseDisco

\textbf{Hardware.} In exact groups, the float64 attention of added heads runs on the GPU up to 2,048 tokens of context and on
the CPU beyond. On CPUs without AVX-512 (\PodsSinAvx{} of the \PodsN{} RunPod L4 machines of the exam), that CPU attention,
rounded to float32, differs from the Google L4 machines in \CpuUno{} or \CpuVeinti{} of \CpuN{} elements, depending on the
model, Gemma or Qwen (relative difference at most $\CpuRel$); the format control does not take this path, and we publish the
machine and CPU of each session's exam parts (amendment E2-X6). For the Leaderboard panel of
\texttt{\XDiecisieteSesion}, whose largest group of added heads makes the whole float64 attention matrix too large for the CPU
library at long context, that attention was computed one GQA group at a time on its longest inputs; on random tokens this gives
logits identical bit for bit to the whole computation wherever the latter fits (amendment E2-X17). The format check of the
Qwen composite, too large for an L4, ran on an H200 against the base model in float32 loaded in the same process (the format
control of that GPU); on two development implants this check gives the same verdict on the H200 and the L4. Its KL part, which
also did not fit in an L4 and had written nothing there, ran on the same H200 from \XVeinteMovida{} CET on 19 November, before
the amendment that lets a part with no output change GPU type was sealed (E2-X20, \XVeinteSellada{} CET). Its panel began on an
L4 GPU and was continued across L4 and H200 GPUs as described next (amendments E2-X20, E2-X21 and E2-X23). \FraseRepartidos{}
\PanelHDoscientosN{} panel instances were computed on H200 GPUs, each compared with a format control computed on the H200 for
exactly that instance (amendment E2-X23); before use, each new H200 had to give base-model records identical to the first one on
a fixed set of inputs, a comparison that first flagged one machine for a difference in a timing field only and was then made to
ignore timing fields, as the one-time certificate does. \FraseDuplicadas{} A panel not finished by \HoraLimite{} is reported as
not measured, never as absence of damage, with its cause where one is known (the slow panels of amendment E2-X21; amendment
E2-X22) and with none otherwise (amendment E2-X27), and the damage prediction of a composite without its panel counts as not
confirmed (E2-X22). An audit of the exam data led to amendment E2-X28 (written at \XVeintiOchoHora{} CET on 19 November,
without seeing any exam result), which changes no decision rule: instances that an older plan had given to a worker no
longer computing them were freed for others; the planning and closing scripts read every readable copy of a machine, not only
the last; a session finished in any archived copy counts as started and is never moved or recomputed; and a RunPod machine is
deleted only after its copy has the same files, bytes and per-instance panel lines in its exam folder as the machine. \FraseAdenda{} \FraseCierre

\begin{table}[h]
\centering
\caption{GPUs of the stage-2 exam. All run the frozen evaluator with the same software stack on a checkpoint with the official
SHA-256. Each kind was checked against the L4 on development implants, never on a sealed set (amendments E2-X6, E2-X9,
E2-X15, E2-X20 and E2-X23), except panel instances on the H200. The H200 was checked before use with Qwen implants; with Gemma
implants, only after its Gemma parts had run. The H100 was checked only with Qwen implants.}
\label{tab:gpus}
\footnotesize
\begin{tabularx}{\linewidth}{@{}l>{\raggedright\arraybackslash}p{3.6cm}>{\raggedright\arraybackslash}X@{}}
\toprule
GPU & Exam parts & Check against the L4 (development implants)\\
\midrule
L4 (Google Cloud) & all & reference\\
L4 (RunPod Secure Cloud) & all & public BBH: \IdenBBH{} of \IdenBBH{} items with identical log-likelihoods; panel: \IdenTray{} of
\IdenTray{} generations identical\\
H100 (RunPod Secure Cloud) & own test and product of the LoRA baseline (B1, all \BUnoCeldas{} cells) & Qwen implants, before
use: 0 of \HCienBBH{} own-test and 0 of \HCienProd{} product predictions differ; never checked with Gemma, whose \BUnoGemmaH{}
B1 cells also ran there\\
H200 (RunPod Secure Cloud) & own test and product; public BBH and KL; parts of the base model and the format control;
format check of the Qwen composite; part of \PanelesHDoscientos{} panels on \HDoscientosMaquinas{} H200 machines, each
instance against an H200 format control & Qwen
implants, before use: 0 of \XNueveBBH{} own-test, 0 of \XNueveProd{} product and 0 of \XQuinceBBH{} public BBH predictions
differ, and KL against the format control is identical (zero on both GPUs for the three implants tested). Panel instances:
never compared with the L4; each was compared only with base-model trajectories computed on the same H200, with the tolerance
calibrated there (E2-X23); H200s give identical base-model records on Qwen inputs.
Gemma implants, after its parts had run: 0 of \XNueveGemmaBBH{}, 0 of \XNueveGemmaProd{} and 0 of \XQuinceGemmaBBH{}; KL against
the format control differs by at most $\XQuinceGemmaKL$ per position, and top-token changes differ in \XQuinceGemmaCambios{} of
\XQuinceGemmaPos{} positions per implant (decision limits: mean KL \HDiezKLMedia{} and \HDiezTopUno{} top-token change per 10,000
tokens)\\
\bottomrule
\end{tabularx}
\end{table}

\end{document}